\documentclass{article}

\usepackage[margin=1in]{geometry}

\usepackage{natbib}

\usepackage[utf8]{inputenc} 
\usepackage[T1]{fontenc}    
\usepackage{hyperref}       
\usepackage{url}            
\usepackage{booktabs}       
\usepackage{amsfonts}       
\usepackage{nicefrac}       
\usepackage{microtype}      
\usepackage{amsthm}
\usepackage{wrapfig}
\usepackage{enumitem}
\usepackage[table]{xcolor}
\usepackage[font=small]{caption}
\usepackage{subcaption}
\usepackage{float}
\usepackage{multirow}
\usepackage{graphicx}
\usepackage{algorithm}
\usepackage{algorithmic}

\usepackage{amsmath,amsfonts,bm}

\def\Tabref#1{Table~\ref{#1}}

\def\Figref#1{Figure~\ref{#1}}

\def\eqref#1{equation~\ref{#1}}

\def\1{\bm{1}}

\def\eps{{\epsilon}}

\def\vzero{{\bm{0}}}

\def\vmu{{\bm{\mu}}}

\def\vphi{{\bm{\phi}}}

\def\veps{{\bm{\epsilon}}}

\def\vb{{\bm{b}}}

\def\vf{{\bm{f}}}
\def\vg{{\bm{g}}}

\def\vk{{\bm{k}}}

\def\vm{{\bm{m}}}

\def\vu{{\bm{u}}}

\def\vw{{\bm{w}}}
\def\vx{{\bm{x}}}
\def\vy{{\bm{y}}}
\def\vz{{\bm{z}}}

\def\mA{{\bm{A}}}

\def\mC{{\bm{C}}}
\def\mD{{\bm{D}}}

\def\mI{{\bm{I}}}

\def\mK{{\bm{K}}}
\def\mL{{\bm{L}}}

\def\mS{{\bm{S}}}

\def\mU{{\bm{U}}}
\def\mV{{\bm{V}}}
\def\mW{{\bm{W}}}
\def\mX{{\bm{X}}}

\def\mZ{{\bm{Z}}}

\def\mPhi{{\bm{\Phi}}}
\def\mLambda{{\bm{\Lambda}}}
\def\mSigma{{\bm{\Sigma}}}

\DeclareMathAlphabet{\mathsfit}{\encodingdefault}{\sfdefault}{m}{sl}
\SetMathAlphabet{\mathsfit}{bold}{\encodingdefault}{\sfdefault}{bx}{n}

\def\gB{{\mathcal{B}}}

\def\gJ{{\mathcal{J}}}

\def\gL{{\mathcal{L}}}

\def\gN{{\mathcal{N}}}
\def\gO{{\mathcal{O}}}

\def\gU{{\mathcal{U}}}

\def\gX{{\mathcal{X}}}

\newcommand{\E}{\mathbb{E}}

\newcommand{\R}{\mathbb{R}}

\newcommand{\KL}{\mathbb{D}_{\mathrm{KL}}}

\newcommand{\Cov}{\mathrm{Cov}}

\DeclareMathOperator{\diag}{diag}

\DeclareMathOperator{\tr}{tr}

\newtheorem{theorem}{Theorem}
\newtheorem{corollary}{Corollary}

\title{Scalable Deep Basis Kernel Gaussian Processes}

\author{%
  Yunqin Zhu$^1$, \quad Henry Shaowu Yuchi$^2$, \quad Yao Xie$^1$\thanks{\texttt{yao.xie@isye.gatech.edu}}\\
  $^1$School of Industrial and Systems Engineering\\
  Georgia Institute of Technology\\
  $^2$Los Alamos National Laboratory\\
}

\begin{document}

\maketitle

\begin{abstract}
  Learning expressive kernels while retaining tractable inference remains a central challenge in scaling Gaussian processes (GPs) to large and complex datasets. We propose a scalable GP regressor based on deep basis kernels (DBKs). Our DBK is constructed from a small set of neural-network-parameterized basis functions with an explicit low-rank structure. This formulation immediately enables linear-complexity inference with respect to the number of samples, possibly without inducing points. DBKs provide a unifying perspective that recovers sparse deep kernel learning and Gaussian Bayesian last-layer methods as special cases. We further identify that naively maximizing the marginal likelihood can lead to oversimplified uncertainty and rank-deficient solutions. To address this, we introduce a mini-batch stochastic objective that directly targets the predictive distribution with decoupled regularization. Empirically, DBKs show advantages in predictive accuracy, uncertainty quantification, and computational efficiency across a range of large-scale regression benchmarks.
\end{abstract}

\section{Introduction}\label{sec:intro}

\begin{figure}[t]
  \centering
  \begin{subfigure}[t]{.62\columnwidth}
    \hspace{.5em}
    \includegraphics[width=0.97\textwidth]{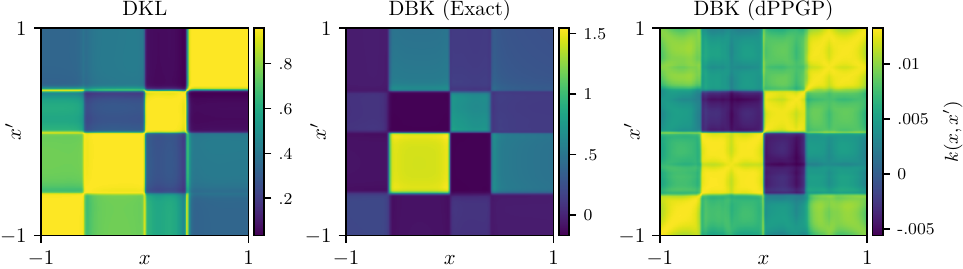}
    \caption{Kernel functions}\vspace{.5em}
  \end{subfigure}\\
  \begin{subfigure}[t]{0.275\columnwidth}
    \centering
    \includegraphics[width=\textwidth]{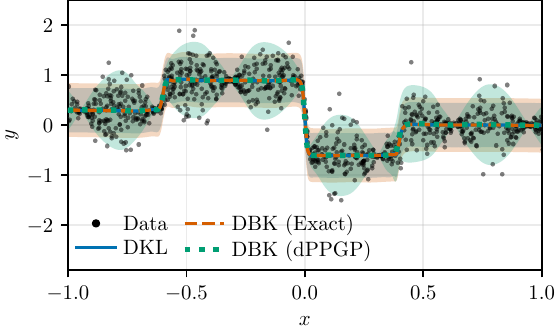}
    \caption{Predictive Mean $\pm$ 2 Std.}
  \end{subfigure}
  \hspace{.1in}
  \begin{subfigure}[t]{0.3\columnwidth}
    \centering
    \raisebox{.4em}{\includegraphics[width=\textwidth]{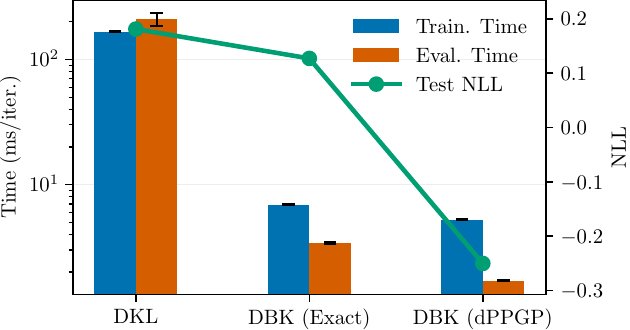}}
    \caption{Performance}
  \end{subfigure}
  \caption{\textbf{Example of fitting a 1-D function.} Data follow a smoothed piecewise-constant mean with sinusoidally varying noise. DBK is the proposed kernel, and dPPGP is the proposed objective. Training time reports one gradient step (full-batch over 20,000 points for exact GPs; mini-batch of size 200 for dPPGP), and evaluation time reports batched prediction over 1,000 test points (shown in (b)). See Section \ref{sec:1d} for details.}
  \label{fig:motivate_exp}
  \vspace{-0.2in}
\end{figure}

Gaussian processes (GPs) provide a principled tool for approximating unknown functions with uncertainty estimates \citep{rasmussenGaussianProcessesMachine2005}. They have been widely applied in scientific modeling, spatial statistics, experimental design, etc., where uncertainty quantification is critical. Central to the expressiveness of a GP is the choice of its covariance function, or \textit{kernel}, which encodes prior assumptions about the function's structure. While classical kernels such as the radial basis function (RBF) and Matern kernels afford analytical convenience and strong inductive biases, they are often too restrictive to capture the non-stationary, non-monotonic patterns present in complex real-world datasets. Moreover, exact GP inference scales cubically with the number of observations, making it prohibitive for massive datasets. This creates a fundamental tension in GP modeling: learning sufficiently expressive kernels while retaining tractable inference.

To overcome the limitations of fixed parametric forms, researchers have pursued more flexible kernels through spectral representations \citep{wilsonGaussianProcessKernels2013,samoGeneralizedSpectralKernels2015,remesNonstationarySpectralKernels2017}, compositional structures \citep{duvenaudStructureDiscoveryNonparametric2013,lloydAutomaticConstructionNaturallanguage2014,sunDifferentiableCompositionalKernel2018}, and neural network (NN) parameterizations \citep{hintonUsingDeepBelief2007,calandraManifoldGaussianProcesses2016,wilsonDeepKernelLearning2016,wilsonStochasticVariationalDeep2016,bradshawAdversarialExamplesUncertainty2017}. Among these, deep kernel learning (DKL) \citep{wilsonDeepKernelLearning2016} integrates NNs into GPs by encoding the inputs through an NN before applying a base kernel in the latent space. DKL leverages the representational power of NNs to learn data-driven kernels and has shown strong empirical performance. 

On the other hand, to reduce the complexity of GP inference, sparse GPs approximate the kernel using a low-rank structure induced by a finite set of inducing points \citep{williamsUsingNystromMethod2000,smolaSparseGreedyMatrix2000,lawrenceFastSparseGaussian2002,snelsonSparseGaussianProcesses2005,quinonero-candelaUnifyingViewSparse2005,titsiasVariationalLearningInducing2009}. This enables scalable inference and mini-batch training, most notably through stochastic variational GPs (SVGPs) \citep{hensmanGaussianProcessesBig2013}, which maximize an ELBO of the marginal likelihood. SVGP has been successfully used to deploy sparse DKL (sDKL) on large datasets \citep{wilsonStochasticVariationalDeep2016}. However, as noted by \citet{jankowiakParametricGaussianProcess2020}, maximum marginal likelihood (MML) training often leads to miscalibrated uncertainty because it mismatches test-time predictions. To address this, a parametric predictive GP (PPGP) objective has been developed that targets the predictive distribution and improves calibration\footnote{Here, calibration refers to predictive calibration (e.g., correct coverage), not marginal-likelihood optimality.} of sDKL.

\begin{figure}[t]
  \centering
  \includegraphics[width=.5\columnwidth]{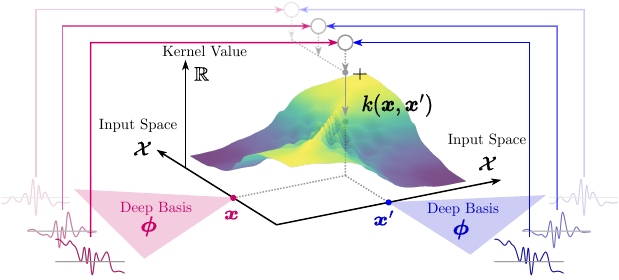}
  \caption{\textbf{Illustration of DBK.} A deep architecture $\vphi$ maps two inputs $\vx$ and $\vx'$ to two sets of basis representations (features) $\{\phi_i(\vx)\}_{i=1}^r$ and $\{\phi_i(\vx')\}_{i=1}^r$. The kernel value $k(\vx,\vx')$ is calculated by summing up the products between two sets of features.}
  \label{fig:dbk}
  \vspace{-0.2in}
\end{figure}

In this work, we introduce scalable GP regressors based on a general family of \textbf{D}eep \textbf{B}asis \textbf{K}ernels (DBKs). DBKs directly represent the kernel as the inner product of a small set of NN-parameterized basis functions, yielding an explicit low-rank structure (\Figref{fig:dbk}). This formulation enables linear-complexity exact inference with respect to the number of training samples. Notably, sparse DKL can be interpreted as a special case of DBKs, in which the basis functions are defined implicitly through inducing points and a whitening transformation \citep{matthewsScalableGaussianProcess2017}. Our DBKs in turn enlarge the design space of deep bases via a two-stage architecture: a neural backbone followed by a flexible expansion layer, possibly without inducing points. This general design also reveals a connection to recently developed Gaussian Bayesian last-layer (GBLL) networks \citep{kristiadiBeingBayesianEven2020,watsonLatentDerivativeBayesian2021,fiedlerImprovedUncertaintyQuantification2023,harrisonVariationalBayesianLast2024}. Specifically, GBLL places a Gaussian prior on the last-layer weights of an NN, equivalent to using the final hidden activations of the network as basis representations. In this sense, DBKs unify sparse sDKL and GBLL within a single low-rank kernel GP framework.

While DBKs provide a flexible and scalable kernel parameterization, the training objective plays a critical role in predictive performance. We observe that naive MML can lead to degenerate rank-1 solutions that oversimplify predictive uncertainty, especially when the data-generating process involves a heteroskedastic noise. Inspired by PPGP \citep{jankowiakParametricGaussianProcess2020}, we propose a novel mini-batch objective, termed dPPGP, that directly targets the predictive distribution with \textbf{d}ecoupled regularization. Specifically, we generalize a common variance correction trick in sparse GPs, which uses an added diagonal term to recover full GP prior variance \citep{quinonero-candelaUnifyingViewSparse2005,titsiasVariationalLearningInducing2009}. Unlike PPGP, we decouple this correction from test-time prediction and impose it instead as a standalone in-batch trace regularizer during training. This allows flexible control over the regularization strength and makes the objective applicable to arbitrary deep bases. The trace regularization encourages a uniform spread of prior uncertainty, acting as an additional mechanism to prevent rank collapse. Empirically, dPPGP achieves improved uncertainty calibration while retaining the scalability of mini-batch optimization.

\Figref{fig:motivate_exp} compares exact DKL, exact DBK, and DBK with the proposed dPPGP objective on 1-D synthetic data. While all methods learn expressive kernels, exact GPs fail to encode sinusoidal signals. DBK (Exact) collapses towards the rank-1 sample covariance up to a constant mean. In contrast, DBK (dPPGP) yields more uniform prior variance, accurately recovers heteroskedastic uncertainty, and achieves the lowest test NLL. The low-rank structure and mini-batch objective substantially reduce inference latency.

This work makes four contributions. (i) We formalize a general low-rank kernel GP regressor with deep bases, which unifies sDKL and GBLL models under a single kernel construction. (ii) We identify and theoretically characterize degenerate solutions that arise from MML training of expressive low-rank kernels. (iii) With these insights, we propose a predictive GP objective with decoupled trace regularization, which improves calibration across both inducing-point and non-inducing-point bases. (iv) Extensive experiments on synthetic and real-world regression data validate that the proposed components enhance predictive accuracy, uncertainty quantification, and computational efficiency.

\section{Preliminaries}\label{sec:prelim}

\paragraph{GPs.} A GP defines a prior over functions $f: \gX \to \R$ on an input space $\gX\subset\R^d$, characterized by a mean function $\mu:\gX \to \R$ and a positive semi-definite covariance function, i.e., kernel $k:\gX \times \gX \to \R$, such that for any finite set of inputs $\mX = \{\vx_i\}_{i=1}^n$, where $\vx_i\in \gX$, the function values are jointly Gaussian distributed as
\begin{equation*}
\begin{gathered}
  \vf = [f(\vx_1), \ldots, f(\vx_n)]^\top \sim \gN(\vmu_\mX, \mK_{\mX\mX}), \\
  \vmu_\mX = [\mu(\vx_1), \ldots, \mu(\vx_n)]^\top,\ \mK_{\mX\mX} = [k(\vx_i, \vx_j)]_{i,j=1}^n.
\end{gathered}
\end{equation*}
In the rest of the paper, we assume $\mu = 0$ w.l.o.g. Assume observations $\vy = \vf + \veps$ with i.i.d. noise $\veps \sim \gN(\vzero, \sigma_\eps^2\mI_n)$. This gives a marginal $\vy \sim \gN(\vzero, \mSigma_{\mX})$ with covariance $\mSigma_{\mX}=\mK_{\mX\mX} + \sigma_\eps^2 \mI_n$. Exact GP training aims to maximize $\log p(\vy)$. For prediction, a closed-form posterior $f(\vx^*)|\vy \sim \gN(\hat\mu_f(\vx^*), \hat\sigma_f^2(\vx^*))$ can be derived:
\begin{equation}\label{eq:posterior}
\begin{aligned}
  \hat\mu_f(\vx^*) & = \vk_\mX(\vx^*)^\top\mSigma_{\mX}^{-1}\vy,\\ \hat\sigma_f^2(\vx^*) & = \ k(\vx^*,\vx^*) - \vk_\mX(\vx^*)^\top\mSigma_{\mX}^{-1}\vk_\mX(\vx^*),
\end{aligned}
\end{equation}
where $\vx^* \in \gX$, $\vk_\mX(\vx^*) = [k(\vx_1, \vx^*), \ldots, k(\vx_n, \vx^*)]^\top$. A major computational bottleneck lies in the inversion of $\mSigma_{\mX}$, which incurs a cost of $\mathcal{O}(n^3)$ time and $\mathcal{O}(n^2)$ space, making standard GP inference impractical for large datasets.

\paragraph{Sparse GPs.} The work by \citet{titsiasVariationalLearningInducing2009} approximates a full GP with kernel $\tilde k$ by a set of $r$ inducing point $\mZ=\{\vz_i\}_{i=1}^r \subset \gX$ and inducing variables $\vu \sim \gN(\vzero, \tilde\mK_{\mZ\mZ})$. An approximate low-rank kernel is constructed:
\begin{equation}\label{eq:sgp}
  k(\vx,\vx') = \tilde\vk^\top_\mZ(\vx) \tilde\mK_{\mZ\mZ}^{-1} \tilde\vk_\mZ(\vx') = \langle \vphi(\vx), \vphi(\vx')\rangle.
\end{equation}
where $\vphi(\vx) = \tilde\mK_{\mZ\mZ}^{-1/2} \tilde\vk_\mZ(\vx)$. This is equivalent to letting $y = \tilde \vk^\top_\mZ(\vx)\tilde\mK_{\mZ\mZ}^{-1} \vu$. Sparse GP training aims to maximize the lower bound (derived in Appendix \ref{sec:trace_derivation}):
\begin{equation}\label{eq:sgp_trace}
  \log \tilde p (\vy) \ge \log p (\vy) - \sum_{\vx} \frac{\tilde k(\vx,\vx) - \|\vphi(\vx)\|^2}{2\sigma_\eps^2},
\end{equation}
where $\tilde p (\vy)$ is the marginal likelihood of the full GP defined by $\tilde k$. Posterior variance is corrected with a diagonal term:
\begin{equation}\label{eq:sgp_corr}
  \tilde{\hat\sigma}_f^2(\vx^*) = \hat\sigma_f^2(\vx^*) + \tilde k(\vx^*,\vx^*) - \|\vphi(\vx^*)\|^2.
\end{equation}
Evaluation of $\log p(\vy)$, $\nabla \log p(\vy)$, $\hat\mu_f(\vx^*)$, $\tilde{\hat\sigma}_f^2(\vx^*)$ takes $\mathcal{O}(nr^2)$ time and $\mathcal{O}(nr)$ space, using low-rank properties. 

\paragraph{SVGP.} Stochastic variational GP (SVGP) proposed by \citet{hensmanGaussianProcessesBig2013} further reduces the cost of a single training iteration to $\mathcal{O}(br^2)$ time and $\mathcal{O}(br)$ memory for batch size $b$. Under the whitening transformation $\vw = \tilde\mK_{\mZ\mZ}^{-1/2}\vu$ \citep{matthewsScalableGaussianProcess2017,jankowiakParametricGaussianProcess2020} and a variational distribution $q(\vw) = \gN(\vw; \vm, \mL\mL^\top)$, the ELBO of $\log p(\vy)$ is 
\begin{equation*}
  \begin{aligned}
  \log p(\vy) \ge &\  \E_{q(\vw)}[\log p(\vy|\vw)] - \KL[q(\vw)\|p(\vw)], \\
  =& \sum_{(\vx, y)} \left\{\log \gN (y; \hat\mu_f(\vx), \sigma_\eps^2) - \frac{\hat\sigma_f^2(\vx)}{2\sigma_\eps^2}\right\} \\
  &- \KL(\gN(\vm, \mL\mL^\top)\|\gN(\vzero,\mI_r)),
  \end{aligned}
\end{equation*}
where $\hat\mu_f(\vx) = \langle\vm, \vphi(\vx)\rangle$, $\hat\sigma_f^2(\vx) = \|\mL^\top \vphi(\vx)\|^2$. Applying (\ref{eq:sgp_trace}) and (\ref{eq:sgp_corr}), the SVGP loss is given by
\begin{equation}\label{eq:svgp}
\begin{aligned}
\gL_{\text{SVGP}}
=&\ \frac{1}{b}\sum_{(\vx, y)} \left\{ 
  - \log \gN (y; \hat\mu_f(\vx), \sigma_\eps^2) + \frac{\tilde{\hat\sigma}_f^2(\vx)}{2\sigma_\eps^2} \right\} \\
&+ \frac{1}{n}\, \KL\left(\gN(\vm, \mL\mL^\top)\,\|\,\gN(0, \mI_r)\right)
\end{aligned}
\end{equation}
The sum is taken over mini-batches. The posterior distribution $\gN(\hat\mu_f(\vx^*), \tilde{\hat\sigma}_f^2(\vx^*))$ can now be evaluated in constant complexity, using the jointly optimized $\vm$ and $\mL$.

\paragraph{PPGP.} The SVGP loss (\ref{eq:svgp}) only involves noise variance $\sigma_\eps^2$ in the data fit term $\log \gN (y; \hat\mu_f(\vx), \sigma_\eps^2)$, yet the full predictive variance consists of both $\tilde{\hat\sigma}_f^2(\vx)$ and $\sigma_\eps^2$. As a result, $\tilde{\hat\sigma}_f^2(\vx)$ is often largely underestimated, with observation noise dominating the uncertainty \citep{jankowiakParametricGaussianProcess2020}. Parametric predictive GP (PPGP) addresses this by directly optimizing the log predictive likelihood with a standalone KL regularizer:
\begin{equation}\label{eq:ppgp}
\begin{aligned}
\gL_{\text{PPGP}}
=&\ \frac{1}{b}\sum_{(\vx, y)} - \log \gN (y; \hat\mu_f(\vx), \tilde{\hat\sigma}_f^2(\vx) + \sigma_\eps^2) \\
&+ \frac{\beta}{n}\, \KL(\gN(\vm, \mL\mL^\top)\,\|\,\gN(0, \mI_r))
\end{aligned}
\end{equation}
where $\beta$ is a hyperparameter that controls the KL regularization strength. While PPGP improves uncertainty quantification compared to exact GP and SVGP, it remains tied to inducing-point kernels and cannot be applied to more general basis representations.

\paragraph{DKL.} Deep kernel learning by \citet{wilsonDeepKernelLearning2016} constructs the kernel as
\begin{equation}\label{eq:dkl}
  k(\vx,\vx') = \tilde k(\vg(\vx), \vg(\vx')),
\end{equation}
where $\vg: \gX \to \R^h$ is an NN and $\tilde k: \R^h \times \R^h \to \R$ is a base kernel, e.g., RBF or spectral mixture \citep{wilsonGaussianProcessKernels2013}, that operates in the $h$-dimensional latent space. While DKL leverages the representational power of NNs, the choice of $\tilde k$ can impose restrictive inductive biases. DKL has been combined with SVGP and PPGP for scalable inference \citep{wilsonStochasticVariationalDeep2016,jankowiakParametricGaussianProcess2020}.

\section{Proposed Method}

\subsection{Deep Basis Kernels}

The discussion in Section \ref{sec:prelim} highlights two recurring themes in modern GP kernel learning: (i) low-rank structure for scalable inference, and (ii) NN parameterization for expressive representation. Deep basis kernels (DBKs) unify these two ideas by constructing kernels directly from a finite set of deep basis functions, yielding an explicit low-rank structure by design. Formally, let $\phi_i:\gX \to \R$, $i=1,\ldots,r$, denote a set of scalar-valued basis functions parameterized by NNs, and define the basis map $\vphi(\vx) = [\phi_1(\vx), \ldots, \phi_r(\vx)]^\top$. Our DBK is given by the inner product:
\begin{equation}\label{eq:dbk}
  k(\vx, \vx') = \sum_{i=1}^r \phi_i(\vx) \phi_i(\vx') = \langle \vphi(\vx), \vphi(\vx') \rangle,
\end{equation}
By construction, $k$ is a valid positive semi-definite function \citep{scholkopf2002learning}. When evaluated at $\mX$, the kernel matrix can be written compactly as $\mK_{\mX\mX} = \mPhi_\mX \mPhi_\mX^\top$, where $\mPhi_\mX \in \R^{n\times r}$ with rows $\{\vphi(\vx_i)^\top\}_{i=1}^n$, and therefore has rank at most $r$. The expressiveness of DBKs is grounded in a classical result on kernel representations. 
\begin{theorem}[Mercer]\label{thm:mercers}
Any continuous, symmetric, positive semi-definite kernel $k: \gX\times \gX \to \R$ on a compact set $\gX\subset \R^d$ admits the expansion \citep{mercerFunctionsPositiveNegative1909a}
\begin{equation}\label{eq:mercer}
  k(\vx, \vx') = \lim_{r\to\infty}\sum_{i=1}^r \lambda_i \psi_i(\vx) \psi_i(\vx'),
\end{equation}
where $\lambda_1 \geq \lambda_2 \geq \dots \geq \lambda_r \geq 0$ are eigenvalues and $\psi_i:\gX\to\R$ are orthonormal eigenfunctions. The series converges absolutely and uniformly on $\gX \times \gX$.
\end{theorem}
Theorem \ref{thm:mercers} implies that any kernel can be truncated to a finite-rank expansion. Moreover, by the universal approximation theorem \citep{hornikMultilayerFeedforwardNetworks1989}, a deep basis can parameterize $\phi_i(\vx) = \sqrt{r}\psi_i(\vx)$ arbitrary well on compact domains. Consequently, DBKs can, in principle, approximate any continuous kernel as the rank $r$ and network capacity increase, provided the eigenvalues decay sufficiently fast.

\begin{figure}[t]
  \centering
  \includegraphics[width=.5\columnwidth]{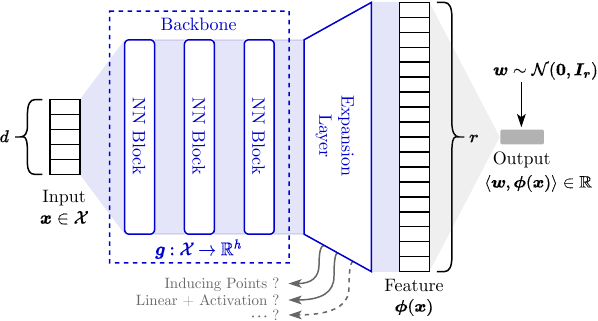}
  \caption{\textbf{Network architecture.} A backbone transforms $\vx$ to an intermediate state, and an expansion layer lifts it to $\vphi(\vx)$. Different choices of expansion layers recover sDKL and GBLL.}
  \label{fig:dbk-nn}
  \vspace{-0.2in}
\end{figure}

\paragraph{Weight-space view.} The kernel (\ref{eq:dbk}) brings up an equivalent weight-space interpretation, which connects low-rank kernel GPs with featurized linear models:
\begin{equation*}
    f(\vx) = \langle \vw, \vphi(\vx) \rangle, \quad \vw \sim \gN(\vzero, \mI_r).
\end{equation*}
In the limit of (\ref{eq:mercer}), this recovers the Karhunen–Loève expansion of a zero-mean GP (Theorem \ref{thm:kl}). This formulation reduces GP inference to Bayesian linear regression in an $r$-dimensional feature space and immediately allows writing the closed-form posterior over weights
\begin{equation}\label{eq:posterior_dbk}
  \vw|\vy \sim \gN\left(\mLambda_\mX^{-1}\mPhi_\mX^\top\vy,\ \sigma_\eps^2\mLambda_\mX^{-1}\right),
\end{equation}
where $\mLambda_\mX = \mPhi_\mX^\top\mPhi_\mX + \sigma_\eps^2\mI_r$. The function posterior in (\ref{eq:posterior}) hence reduces to $\hat\mu_f(\vx^*) = \vphi(\vx^*)^\top\mLambda_\mX^{-1}\mPhi_\mX^\top\vy$ and $\hat\sigma_f^2(\vx^*) = \sigma_\eps^2\vphi(\vx^*)^\top\mLambda_\mX^{-1}\vphi(\vx^*)$. Note that it can be evaluated in $\gO(n(r^2+c_t))$ time and $\gO(n(r+c_s))$ space, where $c_t$, $c_s$ denote the per-input cost of $\vphi$.

\paragraph{MML.} DBKs can be trained exactly by maximizing the log marginal likelihood $\log p(\vy)$ in a scalable form:
\begin{equation}\label{eq:lml_dbk}
\begin{aligned}
  \log p(\vy) = &- \frac{n}{2}\log(2\pi) - \frac{n-r}{2}\log (\sigma_\eps^2)  - \frac{1}{2}\log|\mLambda_\mX| \\
  &- \frac{1}{2\sigma_\eps^2}\|\vy\|_2^2 + \frac{1}{2\sigma_\eps^2}\|\mLambda_\mX^{-1/2}\mPhi_\mX^\top\vy\|_2^2,
\end{aligned}
\end{equation}
The full derivation is provided in Appendix~\ref{sec:exact}. Using the chain rule, the gradient $\nabla\log p(\vy)$ can likewise be computed efficiently. Overall, the low-rank structure reduces the time and space complexity of exact GP inference from $\gO(n^3)$ and $\gO(n^2)$ to $\gO(nr^2)$ and $\gO(nr)$, respectively, yielding significant savings when $r \ll n$.

\paragraph{Design of deep bases.} A key advantage of DBKs is the flexibility in designing deep basis functions. We adopt a general two-stage architecture that separates representation learning from basis expansion, as illustrated in \Figref{fig:dbk-nn}. First, a backbone deep network $\vg: \gX\to \R^h$, e.g., MLP or ResNet, maps the input $\vx$ to an shared $h$-dimensional representation $\vg(\vx)$. Subsequently, a custom expansion layer lifts $\vg(\vx)$ to $r$ basis functions, where the rank $r$ can be chosen independently of $h$. Intuitively, the backbone learns task-relevant abstractions from raw inputs, while the expansion layer explicitly controls the rank and determines the fundamental structure of the basis functions. Depending on the choice of expansion, this architecture subsumes several closely related methods as special cases:
\begin{itemize}
    \item \textbf{sDKL} \citep{wilsonDeepKernelLearning2016,wilsonStochasticVariationalDeep2016}. Combining the sparse GP approximation in (\ref{eq:sgp}) with the deep kernel in (\ref{eq:dkl}) yields basis functions of the form 
    \begin{equation}\label{eq:case_sdkl}
    \vphi(\vx) = \tilde\mK_{\mZ\mZ}^{-1/2} \tilde\vk_\mZ(\vg(\vx)),
    \end{equation}
    where $\mZ=\{\vz_i\}_{i=1}^r \subset \R^h$ are learnable inducing points in the latent space, and $\tilde k$ is a base kernel with additional parameters, e.g., lengthscales.
    \item \textbf{GBLL} \citep{kristiadiBeingBayesianEven2020,watsonLatentDerivativeBayesian2021,fiedlerImprovedUncertaintyQuantification2023,harrisonVariationalBayesianLast2024}. These models adapt vanilla NNs by placing a Gaussian prior on the last-layer weights $\vw$, effectively treating the penultimate-layer activations as $\vphi$:
    \begin{equation}\label{eq:case_gbll}
    \vphi(\vx) = a(\mW\vg(\vx) + \vb),
    \end{equation}
    where $\mW$, $\vb$ are the parameters of a dense linear layer, and $a(\cdot)$ is a point-wise activation function.
\end{itemize}
To conclude, DBKs provide a unifying perspective of low-rank deep GP kernels. Beyond the aforementioned cases, other expansion layers, such as Fourier features \citep{rahimiRandomFeaturesLargescale2007,lazaro-gredillaSparseSpectrumGaussian2010} and B-splines \citep{cunninghamActuallySparseVariational2023}, can also be accommodated. A systematic study of this design space is left to future work.

\subsection{dPPGP}

The learning objective is critical in shaping uncertainty estimates of GP regressors. Particularly, we identify that MML can be ill-suited for use with flexible low-rank kernels.

Assume that we have fixed inputs $\mX$ and regression targets $\vy$ from a true data-generating distribution $p_{\rm gt}(\vy) = \gN(\vy; \vmu_{\rm gt}, \mD_{\rm gt})$, where $\vmu_{\rm gt} = [\mu_{\rm gt}(\vx_1),\dots,\mu_{\rm gt}(\vx_n)]^\top$, $\mD_{\rm gt} = \diag(\sigma_{\rm gt}^2(\vx_1),\ldots,\sigma_{\rm gt}^2(\vx_n))$, $\mu_{\rm gt}: \gX\to \R$, $\sigma^2_{\rm gt}: \gX\to \R^+$ are the ground-truth mean and variance functions, respectively. Then, the oracle of marginal covariance $\mSigma_\mX$ under MML can be characterized by:
\begin{theorem}[Oracle covariance]\label{thm:oracle_Sigma} The optimal covariance $\mSigma^*_\mX$ maximizing the expected log marginal likelihood $\E_{p_{\rm gt}(\vy)}[\log p(\vy)]$ is the sample covariance
\begin{equation}\label{eq:Sgt_def}
  \mSigma^*_\mX = \mS_\mX := \E_{p_{\rm gt}(\vy)}[\vy\vy^\top] = \vmu_{\rm gt}\vmu_{\rm gt}^\top + \mD_{\rm gt}.
\end{equation}
\end{theorem}

\begin{corollary}[Rank-$1$ degeneracy, homogeneous $\sigma^2_{\rm gt}$]\label{cor:rank1_homo}
If $\mD_{\rm gt}=\sigma_{\rm gt}^2\mI_n$, then under MML, the optimal kernel matrix and noise variance are $\mK_{\mX\mX}^* = \vf_{\rm gt}\vf_{\rm gt}^\top$, $\sigma^{*2}_\eps = \sigma^2_{\rm gt}$, respectively. In other words, the solution is rank-$1$.
\end{corollary}

Although a rank-$1$ solution can perfectly fit the ground-truth mean, it collapses the whole function space onto the same direction. The resulting posterior variance $\hat\sigma_f^2$ only measures the signal amplitude and does not capture richer structures for uncertainty. For the general heteroskedastic case, MML leads to a different but still problematic behavior:

\begin{theorem}[Low-rank solution, heteroskedastic $\sigma^2_{\rm gt}$]\label{thm:hetero} Let the eigendecomposition of the oracle covariance (\ref{eq:Sgt_def}) be $\mS_\mX  = \mU\diag(\lambda_1,\ldots,\lambda_n)\mU^\top$, $\lambda_1\ge \cdots \ge \lambda_n \ge 0$. Under MLL, the optimal rank-$r$ kernel matrix and noise variance are $\mK_{\mX\mX}^* = \sum_{i=1}^r (\lambda_i-\sigma_{\eps}^{*2}) \vu_i\vu_i^\top$ and $\sigma_{\eps}^{*2} = \frac{1}{n-r}\sum_{i=r+1}^n \lambda_i$, respectively, where $\{\lambda_i\}_{i=r+1}^n$ interlace with the diagonal entries of $\mD_{\rm gt}$.
\end{theorem}

In this solution, the model noise variance $\sigma_{\eps}^2$ can be viewed as a scalar average of the ground-truth $\sigma^2_{\rm gt}$ across the input space. This means that the model amortizes the heteroskedastic noise via a mis-specified homogeneous component $\sigma_{\eps}^2$, rather than estimating it with the input-dependent function variance $\hat\sigma_f^2$. As a result, the full predictive variance $(\hat\sigma_f^2 + \sigma^2_\eps)$ is largely insensitive to $\vx$, leading to overconfident predictions in high-noise regions and vice versa.

To address the issues, we generalize ideas from sparse GPs and introduce an improved mini-batch objective, termed dPPGP. Given a variational (pseudo-)posterior distribution $q(\vw)=\gN(\vw;\vm,\mL\mL^\top)$, the predictive mean and function variance of a DBK GP are written as
\begin{equation*}
\hat\mu_f(\vx)=\langle\vm,\vphi(\vx)\rangle,\quad
\hat\sigma_f^2(\vx)=\|\mL^\top\vphi(\vx)\|^2.
\end{equation*}
Following PPGP, we define the objective using the log predictive likelihood $\log\gN (y;\hat\mu_f(\vx),\hat\sigma_f^2(\vx)+\sigma_\eps^2)$ as the data fit term. This formulation forces the full predictive variance to explain input-dependent uncertainty, thereby aligning training more closely with test-time prediction.

A key caveat is that PPGP, as a sparse GP objective, relies on an explicit variance correction derived from the base kernel definition (\ref{eq:sgp_corr}), which is not available for a general basis map $\vphi$. To this end, we decouple the correction from test time and instead impose it as a regularizer during training:
\begin{equation*}
    \gL_{\text{trace}} = \frac{1}{b} \sum_{\vx} \frac{\tilde{k}_b - \|\vphi(\vx)\|^2}{2\sigma_\eps^2},\ \tilde{k}_b = \max_{\vx} \|\vphi(\vx)\|^2
\end{equation*}
where both the sum and maximum are taken over a batch of size $b$. This regularizer is inherited from (\ref{eq:sgp_trace}), with the distinction that the exact kernel diagonal $\tilde k(\vx,\vx)$ is replaced by an in-batch proxy $\tilde{k}_b$, making it applicable to arbitrary deep bases. We refer to this as {\it trace regularization}, as it is related to the trace of a PSD error matrix (see Appendix \ref{sec:trace_derivation}). Intuitively, minimizing $\gL_{\text{trace}}$ encourages a more uniform allocation of prior variance $\|\vphi(\vx)\|^2$ across inputs, providing an additional mechanism to prevent rank collapse. 

The complete dPPGP loss is given by
\begin{equation}\label{eq:dppgp}
\begin{aligned}
\gL_{\text{dPPGP}}
= &\ \frac{1}{b}\sum_{(\vx, y)} - \log \gN (y; \hat\mu_f(\vx),\hat\sigma_f^2(\vx) + \sigma_\eps^2) \\
  &+ \alpha\, \gL_{\text{trace}} + \frac{\beta}{n}\, \KL(\gN(\vm, \mL\mL^\top)\,\|\,\gN(0, \mI_r)),
\end{aligned}
\end{equation}
where $\alpha$ and $\beta$ control the strength of the trace regularization and KL regularization, respectively.

\section{Experiments}\label{sec:experiments}
We investigate the empirical performance of DBK GPs across a series of large-scale regression tasks, including a 1-D example, UCI benchmarks, and an application of internet quality estimation. All NN-based models share a ResNet \citep{he2016deep,gorishniyRevisitingDeepLearning2023} backbone $\vg(\cdot)$ with SiLU activation. For standard GPs and DKL, an RBF (base) kernel is used. As per the choice of expansion layers, two DBK variants are considered: (i) \textbf{DBK-SiLU}, using a fully connected layer with SiLU activation for basis expansion, i.e., the GBLL case (\ref{eq:case_gbll}); (ii) \textbf{DBK-RBF}, using an RBF inducing-point kernel for basis expansion, i.e., the sDKL case (\ref{eq:case_sdkl}). A full description of the baselines and detailed settings is available in Appendix \ref{sec:exp_details}.

\begin{figure*}[t]
  \centering
  \includegraphics[width=\textwidth]{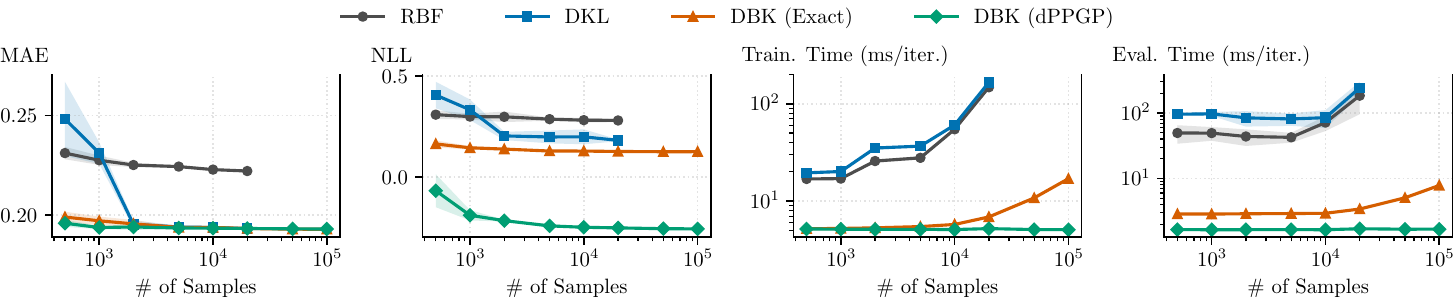}
  \caption{\textbf{Scaling behavior on 1-D synthetic data.} We report test MAE and NLL (lower is better). Training time measures a single gradient step (full-batch for RBF/DKL/DBK (Exact); mini-batch of size 200 for DBK (dPPGP)). Evaluation time measures a single batched prediction over all 1,000 test points. Due to memory constraints, exact RBF and DKL GPs do not scale beyond 20,000 training samples. Shaded regions indicate one standard deviation over 10 random seeds.}
  \label{fig:step1d_scaling}
  \vspace{-0.05in}
\end{figure*}

\begin{figure*}[t]
  \centering
  \begin{subfigure}[t]{0.7\textwidth}
    \centering
    \raisebox{.7em}{\includegraphics[width=\textwidth]{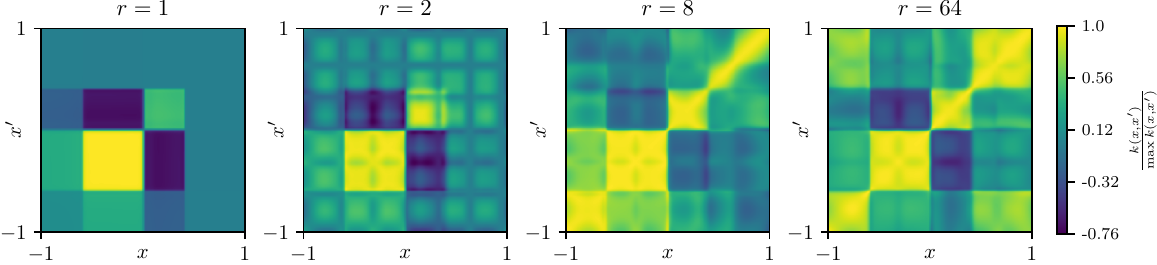}}
  \end{subfigure}
  \begin{subfigure}[t]{0.28\textwidth}
    \includegraphics[width=\textwidth]{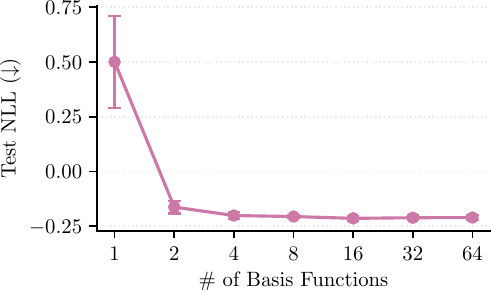}
  \end{subfigure}
  \caption{\textbf{Effect of rank on 1-D synthetic data.} Left: Learned DBKs of rank $r \in \{1,2,8,64\}$, each normalized by its maximum value on the input grid. Right: Test negative log-likelihood versus rank. Error bars indicate one standard deviation over 10 random seeds. Additional visualizations of the learned basis functions and predictive distributions are provided in \Figref{fig:rank_viz}.}
  \label{fig:1d_rank}
  \vspace{-0.1in}
\end{figure*}

\subsection{1-D Synthetic Data}\label{sec:1d}
We consider a 1-D regression dataset generated by randomly sampling inputs $x_i$ from $\gU([-1,1])$ and targets $y_i$ from $\gN(y_i; \mu_{\rm gt}(x_i), \sigma^2_{\rm gt}(x_i))$, where $\mu_{\rm gt}(x)$ is constructed via sharp transitions $\mu_{\rm gt}(x)=0.3(1-s_1)+0.9(s_1-s_2)-0.6(s_2-s_3)$ with $s_1=\varsigma(200(x+0.6))$, $s_2=\varsigma(200x)$, $s_3=\varsigma(200(x-0.4))$, and $\varsigma(\cdot)$ denotes the logistic function. The variance is set to $\sigma_{\rm gt}^2(x)=(2\sin(10x))^2$, yielding input-dependent uncertainty. We focus on the DBK-SiLU variant in this example, using either exact inference or dPPGP, and compare it against exact RBF and DKL. 

To show the scaling behavior of these GP regressors, we vary $n$ from $500$ to $100,000$ and report the corresponding results in \Figref{fig:step1d_scaling}. DBK performs consistently the best in both accuracy and calibration, with NLL substantially improved by dPPGP. This shows the strength of non-inducing deep bases in capturing sharp, step-like transitions that RBFs may otherwise over-smooth, as well as the importance of aligning the training objective with the test-time predictive distribution. In addition, exact DBK scales to $n=10^5$ with dramatically lower latency due to linear complexity. The efficiency is further improved by mini-batch training and a variational formulation in dPPGP, highlighting its practical advantages in large-sample regimes.

Furthermore, we investigate the effect of rank by varying $r$ from 1 to 64 and visualize the learned DBK in \Figref{fig:1d_rank} with the test NLL. For very small ranks $r=1,2$, the kernel is overly restrictive, leading to high NLL. In particular, when $r=1$, the model learns a single global feature (\Figref{fig:rank_viz}) that resembles $\mu_{\rm gt}$, resulting in a blocky kernel structure and failure to capture local variations. As $r$ increases, the kernel becomes more expressive and better encodes the sinusoidally varying uncertainty, while the performance gains saturate after a moderate rank $r\approx8$, indicating that relatively low ranks suffice to achieve calibrated predictions.

\subsection{UCI Regression}

Next, we benchmark two DBK variants with either inducing-point or non-inducing-point bases against baselines on 10 UCI univariate regression datasets of varying scales \footnote{\url{https://archive.ics.uci.edu}}. Dataset sizes $(n,d)$ are shown in the header of \Tabref{tab:uci}. For baselines, we include a plain deep neural network (\textbf{DNN}) trained with MSE, and three closely related neural GP methods: (i) \textbf{VBLL} \citep{harrisonVariationalBayesianLast2024}, which maximizes the ELBO of a GBLL; (ii) \textbf{SV-DKL} \citep{wilsonStochasticVariationalDeep2016}, which approximates DKL with inducing points under SVGP; (iii) \textbf{PP-DKL} \citep{jankowiakParametricGaussianProcess2020}, which approximates DKL with inducing points under PPGP. For both DBK-SiLU and DBK-RBF, we use the proposed dPPGP. All models support mini-batch training and, except for DNN, allow sampling-free, single-pass uncertainty quantification.

\begin{wrapfigure}{r}{.48\textwidth}
  \centering
  \begin{subfigure}[t]{0.23\textwidth}
    \includegraphics[width=\textwidth]{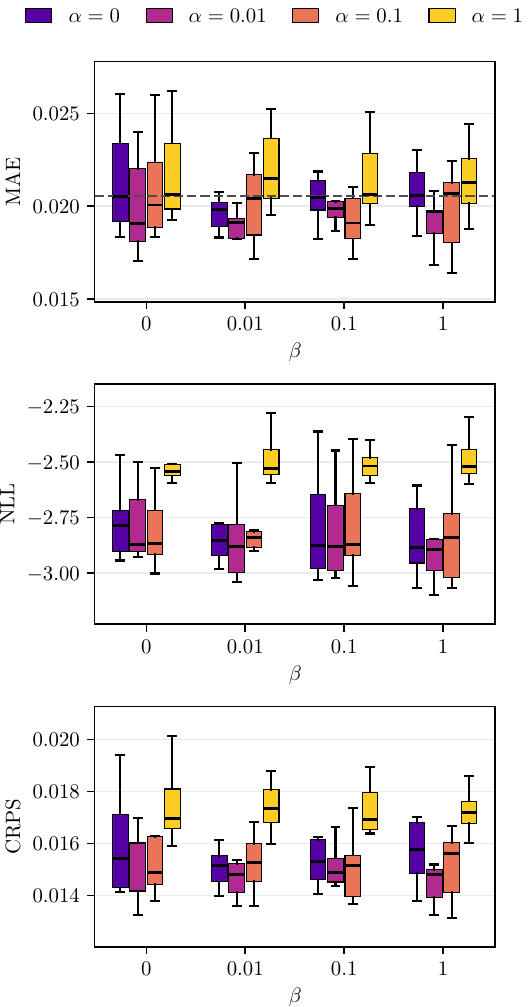}
    \caption{Pol}
  \end{subfigure}
  \begin{subfigure}[t]{0.23\textwidth}
    \centering
    \includegraphics[width=\textwidth]{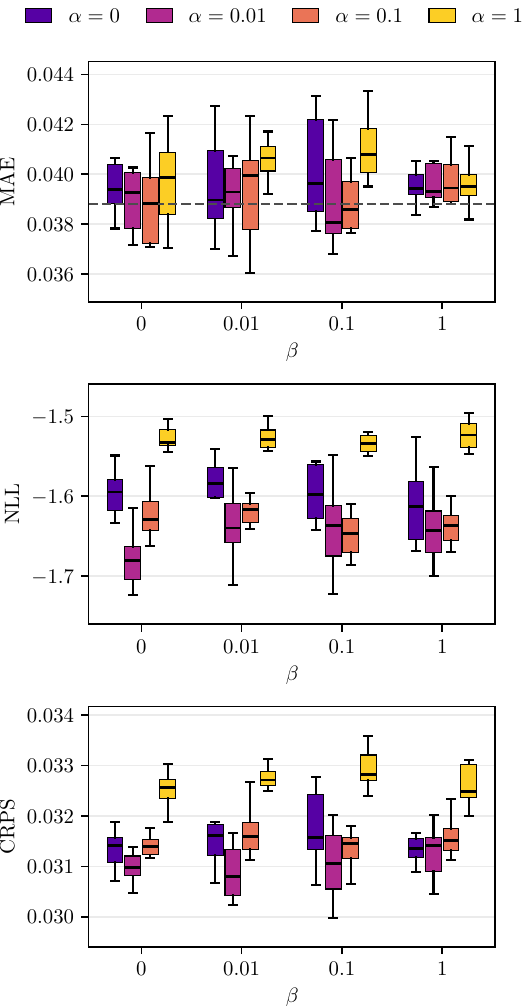}
    \caption{Keggdir.}
  \end{subfigure}
  \caption{\textbf{Sensitivity to $\alpha$ and $\beta$ on UCI regression.} We show box plots of the test metrics (lower is better) of DBK-SiLU over 10 random seeds. Dashed lines indicate the VBLL baseline.}
  \label{fig:sensitivity}
\end{wrapfigure}

\begin{table}[t]
\centering
\setlength{\tabcolsep}{1pt}
\renewcommand{\arraystretch}{1.2}
\caption{\textbf{UCI regression benchmarks.} We report test performance as mean $\pm$ standard deviation over 5 random seeds. For each dataset and metric, the best result is shown in \textcolor{blue!70!black}{\textbf{bold}} and the second-best is \textcolor{blue!55!black}{\underline{underlined}} (lower is better). We also report ($n$, $d$) for each dataset, where $n$ is the number of training samples and $d$ is the input dimension. The rightmost column shows the average rank of each method across all datasets for the corresponding metric. The proposed DBK variants are highlighted in light blue.}\label{tab:uci}
\resizebox{\textwidth}{!}{%
\begin{tabular}{@{}ll|cccccccccc|c@{}}
\toprule
\multirow{2}{*}{Metric} & \multirow{2}{*}{Method} & Pol & Elevators & Kin40K & Protein & Keggdir. & Slice & Keggundir. & 3Droad & Song & Buzz & \multirow{2}{*}{Rank} \\
 &  & (12000,26) & (13279,18) & (32000,8) & (36584,9) & (39061,20) & (42800,385) & (50886,27) & (347898,3) & (412275,90) & (466600,77) &  \\
\midrule
\midrule
\multirow[t]{6}{*}{MAE} & DNN & .0222 {\textcolor{gray}{\small $\pm$ .0009}} & .2629 {\textcolor{gray}{\small $\pm$ .0028}} & \textcolor{blue!70!black}{\textbf{.0250 {\textcolor{gray}{\small $\pm$ .0004}}}} & .3853 {\textcolor{gray}{\small $\pm$ .0042}} & .0392 {\textcolor{gray}{\small $\pm$ .0015}} & \textcolor{blue!70!black}{\textbf{.0109 {\textcolor{gray}{\small $\pm$ .0002}}}} & .0461 {\textcolor{gray}{\small $\pm$ .0011}} & .2435 {\textcolor{gray}{\small $\pm$ .0048}} & .6072 {\textcolor{gray}{\small $\pm$ .0007}} & \textcolor{blue!70!black}{\textbf{.1685 {\textcolor{gray}{\small $\pm$ .0002}}}} & 3.70 \\
 & VBLL & .0205 {\textcolor{gray}{\small $\pm$ .0014}} & \textcolor{blue!70!black}{\textbf{.2605 {\textcolor{gray}{\small $\pm$ .0022}}}} & \textcolor{blue!55!black}{\underline{.0268 {\textcolor{gray}{\small $\pm$ .0013}}}} & \textcolor{blue!55!black}{\underline{.3825 {\textcolor{gray}{\small $\pm$ .0038}}}} & .0388 {\textcolor{gray}{\small $\pm$ .0009}} & .0133 {\textcolor{gray}{\small $\pm$ .0028}} & \textcolor{blue!70!black}{\textbf{.0447 {\textcolor{gray}{\small $\pm$ .0006}}}} & \textcolor{blue!70!black}{\textbf{.2083 {\textcolor{gray}{\small $\pm$ .0029}}}} & .6067 {\textcolor{gray}{\small $\pm$ .0008}} & \textcolor{blue!55!black}{\underline{.1690 {\textcolor{gray}{\small $\pm$ .0003}}}} & \textcolor{blue!70!black}{\textbf{2.60}} \\
 & SV-DKL & .0212 {\textcolor{gray}{\small $\pm$ .0015}} & .2636 {\textcolor{gray}{\small $\pm$ .0049}} & .0275 {\textcolor{gray}{\small $\pm$ .0017}} & \textcolor{blue!70!black}{\textbf{.3703 {\textcolor{gray}{\small $\pm$ .0041}}}} & .0391 {\textcolor{gray}{\small $\pm$ .0014}} & .0139 {\textcolor{gray}{\small $\pm$ .0023}} & \textcolor{blue!55!black}{\underline{.0447 {\textcolor{gray}{\small $\pm$ .0011}}}} & .2324 {\textcolor{gray}{\small $\pm$ .0043}} & .6062 {\textcolor{gray}{\small $\pm$ .0010}} & .1692 {\textcolor{gray}{\small $\pm$ .0001}} & 3.90 \\
 & PP-DKL & .0205 {\textcolor{gray}{\small $\pm$ .0013}} & .2614 {\textcolor{gray}{\small $\pm$ .0020}} & .0296 {\textcolor{gray}{\small $\pm$ .0012}} & .4071 {\textcolor{gray}{\small $\pm$ .0125}} & .0392 {\textcolor{gray}{\small $\pm$ .0018}} & .0123 {\textcolor{gray}{\small $\pm$ .0010}} & .0463 {\textcolor{gray}{\small $\pm$ .0025}} & .2527 {\textcolor{gray}{\small $\pm$ .0037}} & \textcolor{blue!55!black}{\underline{.6034 {\textcolor{gray}{\small $\pm$ .0028}}}} & .1693 {\textcolor{gray}{\small $\pm$ .0005}} & 4.70 \\
\rowcolor{blue!6} & 
DBK-SiLU & \textcolor{blue!55!black}{\underline{.0190 {\textcolor{gray}{\small $\pm$ .0011}}}} & \textcolor{blue!55!black}{\underline{.2606 {\textcolor{gray}{\small $\pm$ .0025}}}} & .0272 {\textcolor{gray}{\small $\pm$ .0008}} & .4031 {\textcolor{gray}{\small $\pm$ .0019}} & \textcolor{blue!55!black}{\underline{.0386 {\textcolor{gray}{\small $\pm$ .0018}}}} & \textcolor{blue!55!black}{\underline{.0122 {\textcolor{gray}{\small $\pm$ .0009}}}} & .0464 {\textcolor{gray}{\small $\pm$ .0025}} & \textcolor{blue!55!black}{\underline{.2285 {\textcolor{gray}{\small $\pm$ .0088}}}} & \textcolor{blue!70!black}{\textbf{.6029 {\textcolor{gray}{\small $\pm$ .0004}}}} & .1692 {\textcolor{gray}{\small $\pm$ .0003}} & \textcolor{blue!55!black}{\underline{3.00}} \\
\rowcolor{blue!6} & 
DBK-RBF & \textcolor{blue!70!black}{\textbf{.0190 {\textcolor{gray}{\small $\pm$ .0010}}}} & .2614 {\textcolor{gray}{\small $\pm$ .0024}} & .0290 {\textcolor{gray}{\small $\pm$ .0017}} & .3937 {\textcolor{gray}{\small $\pm$ .0023}} & \textcolor{blue!70!black}{\textbf{.0382 {\textcolor{gray}{\small $\pm$ .0010}}}} & .0123 {\textcolor{gray}{\small $\pm$ .0020}} & .0451 {\textcolor{gray}{\small $\pm$ .0024}} & .2423 {\textcolor{gray}{\small $\pm$ .0039}} & .6036 {\textcolor{gray}{\small $\pm$ .0015}} & .1690 {\textcolor{gray}{\small $\pm$ .0003}} & 3.10 \\
\midrule
\multirow[t]{6}{*}{NLL} & DNN & -- & -- & -- & -- & -- & -- & -- & -- & -- & -- & -- \\
 & VBLL & -1.6801 {\textcolor{gray}{\small $\pm$ .0168}} & .3623 {\textcolor{gray}{\small $\pm$ .0080}} & -1.8664 {\textcolor{gray}{\small $\pm$ .0420}} & .8231 {\textcolor{gray}{\small $\pm$ .0048}} & -1.0509 {\textcolor{gray}{\small $\pm$ .0056}} & -2.1368 {\textcolor{gray}{\small $\pm$ .3262}} & -.6915 {\textcolor{gray}{\small $\pm$ .0020}} & .2193 {\textcolor{gray}{\small $\pm$ .0112}} & 1.1732 {\textcolor{gray}{\small $\pm$ .0016}} & -.0231 {\textcolor{gray}{\small $\pm$ .0009}} & 4.30 \\
 & SV-DKL & -1.7034 {\textcolor{gray}{\small $\pm$ .0209}} & .3607 {\textcolor{gray}{\small $\pm$ .0115}} & -1.8244 {\textcolor{gray}{\small $\pm$ .0370}} & .8262 {\textcolor{gray}{\small $\pm$ .0100}} & -1.0407 {\textcolor{gray}{\small $\pm$ .0141}} & -2.2643 {\textcolor{gray}{\small $\pm$ .3481}} & -.6870 {\textcolor{gray}{\small $\pm$ .0025}} & .3276 {\textcolor{gray}{\small $\pm$ .0207}} & 1.1737 {\textcolor{gray}{\small $\pm$ .0012}} & -.0215 {\textcolor{gray}{\small $\pm$ .0005}} & 4.70 \\
 & PP-DKL & -2.9551 {\textcolor{gray}{\small $\pm$ .0851}} & .2989 {\textcolor{gray}{\small $\pm$ .0054}} & -2.1023 {\textcolor{gray}{\small $\pm$ .0231}} & .7445 {\textcolor{gray}{\small $\pm$ .0182}} & \textcolor{blue!55!black}{\underline{-1.6548 {\textcolor{gray}{\small $\pm$ .0259}}}} & \textcolor{blue!55!black}{\underline{-2.4330 {\textcolor{gray}{\small $\pm$ .1376}}}} & -1.7395 {\textcolor{gray}{\small $\pm$ .0973}} & -.0801 {\textcolor{gray}{\small $\pm$ .0151}} & 1.1155 {\textcolor{gray}{\small $\pm$ .0010}} & \textcolor{blue!55!black}{\underline{-.1187 {\textcolor{gray}{\small $\pm$ .0015}}}} & 2.70 \\
\rowcolor{blue!6} & 
DBK-SiLU & \textcolor{blue!55!black}{\underline{-2.9670 {\textcolor{gray}{\small $\pm$ .0550}}}} & \textcolor{blue!70!black}{\textbf{.2936 {\textcolor{gray}{\small $\pm$ .0060}}}} & \textcolor{blue!70!black}{\textbf{-2.1554 {\textcolor{gray}{\small $\pm$ .0334}}}} & \textcolor{blue!70!black}{\textbf{.7406 {\textcolor{gray}{\small $\pm$ .0200}}}} & -1.6479 {\textcolor{gray}{\small $\pm$ .0322}} & -2.3533 {\textcolor{gray}{\small $\pm$ .3355}} & \textcolor{blue!55!black}{\underline{-1.7551 {\textcolor{gray}{\small $\pm$ .0795}}}} & \textcolor{blue!70!black}{\textbf{-.1978 {\textcolor{gray}{\small $\pm$ .0320}}}} & \textcolor{blue!70!black}{\textbf{1.1126 {\textcolor{gray}{\small $\pm$ .0010}}}} & -.1181 {\textcolor{gray}{\small $\pm$ .0020}} & \textcolor{blue!55!black}{\underline{1.80}} \\
\rowcolor{blue!6} & 
DBK-RBF & \textcolor{blue!70!black}{\textbf{-2.9807 {\textcolor{gray}{\small $\pm$ .0648}}}} & \textcolor{blue!55!black}{\underline{.2967 {\textcolor{gray}{\small $\pm$ .0137}}}} & \textcolor{blue!55!black}{\underline{-2.1120 {\textcolor{gray}{\small $\pm$ .0337}}}} & \textcolor{blue!55!black}{\underline{.7416 {\textcolor{gray}{\small $\pm$ .0081}}}} & \textcolor{blue!70!black}{\textbf{-1.6679 {\textcolor{gray}{\small $\pm$ .0423}}}} & \textcolor{blue!70!black}{\textbf{-2.4364 {\textcolor{gray}{\small $\pm$ .2063}}}} & \textcolor{blue!70!black}{\textbf{-1.7755 {\textcolor{gray}{\small $\pm$ .1277}}}} & \textcolor{blue!55!black}{\underline{-.1212 {\textcolor{gray}{\small $\pm$ .0198}}}} & \textcolor{blue!55!black}{\underline{1.1136 {\textcolor{gray}{\small $\pm$ .0013}}}} & \textcolor{blue!70!black}{\textbf{-.1202 {\textcolor{gray}{\small $\pm$ .0014}}}} & \textcolor{blue!70!black}{\textbf{1.50}} \\
\midrule
\multirow[t]{6}{*}{CRPS} & DNN & -- & -- & -- & -- & -- & -- & -- & -- & -- & -- & -- \\
 & VBLL & .0215 {\textcolor{gray}{\small $\pm$ .0005}} & .1896 {\textcolor{gray}{\small $\pm$ .0012}} & \textcolor{blue!55!black}{\underline{.0198 {\textcolor{gray}{\small $\pm$ .0009}}}} & .2884 {\textcolor{gray}{\small $\pm$ .0023}} & .0364 {\textcolor{gray}{\small $\pm$ .0004}} & .0099 {\textcolor{gray}{\small $\pm$ .0021}} & .0508 {\textcolor{gray}{\small $\pm$ .0002}} & \textcolor{blue!55!black}{\underline{.1608 {\textcolor{gray}{\small $\pm$ .0018}}}} & .4354 {\textcolor{gray}{\small $\pm$ .0006}} & .1265 {\textcolor{gray}{\small $\pm$ .0001}} & 4.00 \\
 & SV-DKL & .0215 {\textcolor{gray}{\small $\pm$ .0007}} & .1911 {\textcolor{gray}{\small $\pm$ .0027}} & .0206 {\textcolor{gray}{\small $\pm$ .0010}} & \textcolor{blue!55!black}{\underline{.2840 {\textcolor{gray}{\small $\pm$ .0027}}}} & .0371 {\textcolor{gray}{\small $\pm$ .0006}} & .0103 {\textcolor{gray}{\small $\pm$ .0017}} & .0507 {\textcolor{gray}{\small $\pm$ .0002}} & .1792 {\textcolor{gray}{\small $\pm$ .0033}} & .4354 {\textcolor{gray}{\small $\pm$ .0006}} & .1267 {\textcolor{gray}{\small $\pm$ .0000}} & 4.20 \\
 & PP-DKL & .0152 {\textcolor{gray}{\small $\pm$ .0008}} & .1873 {\textcolor{gray}{\small $\pm$ .0009}} & .0212 {\textcolor{gray}{\small $\pm$ .0006}} & .2875 {\textcolor{gray}{\small $\pm$ .0078}} & .0317 {\textcolor{gray}{\small $\pm$ .0005}} & .0091 {\textcolor{gray}{\small $\pm$ .0007}} & .0380 {\textcolor{gray}{\small $\pm$ .0015}} & .1766 {\textcolor{gray}{\small $\pm$ .0025}} & .4259 {\textcolor{gray}{\small $\pm$ .0015}} & \textcolor{blue!55!black}{\underline{.1230 {\textcolor{gray}{\small $\pm$ .0003}}}} & 3.30 \\
\rowcolor{blue!6} & 
DBK-SiLU & \textcolor{blue!55!black}{\underline{.0144 {\textcolor{gray}{\small $\pm$ .0008}}}} & \textcolor{blue!70!black}{\textbf{.1865 {\textcolor{gray}{\small $\pm$ .0014}}}} & \textcolor{blue!70!black}{\textbf{.0195 {\textcolor{gray}{\small $\pm$ .0006}}}} & .2852 {\textcolor{gray}{\small $\pm$ .0015}} & \textcolor{blue!70!black}{\textbf{.0308 {\textcolor{gray}{\small $\pm$ .0005}}}} & \textcolor{blue!70!black}{\textbf{.0090 {\textcolor{gray}{\small $\pm$ .0006}}}} & \textcolor{blue!55!black}{\underline{.0376 {\textcolor{gray}{\small $\pm$ .0008}}}} & \textcolor{blue!70!black}{\textbf{.1599 {\textcolor{gray}{\small $\pm$ .0059}}}} & \textcolor{blue!70!black}{\textbf{.4253 {\textcolor{gray}{\small $\pm$ .0002}}}} & .1230 {\textcolor{gray}{\small $\pm$ .0002}} & \textcolor{blue!70!black}{\textbf{1.60}} \\
\rowcolor{blue!6} & 
DBK-RBF & \textcolor{blue!70!black}{\textbf{.0144 {\textcolor{gray}{\small $\pm$ .0006}}}} & \textcolor{blue!55!black}{\underline{.1870 {\textcolor{gray}{\small $\pm$ .0014}}}} & .0209 {\textcolor{gray}{\small $\pm$ .0012}} & \textcolor{blue!70!black}{\textbf{.2817 {\textcolor{gray}{\small $\pm$ .0019}}}} & \textcolor{blue!55!black}{\underline{.0311 {\textcolor{gray}{\small $\pm$ .0007}}}} & \textcolor{blue!55!black}{\underline{.0091 {\textcolor{gray}{\small $\pm$ .0015}}}} & \textcolor{blue!70!black}{\textbf{.0368 {\textcolor{gray}{\small $\pm$ .0012}}}} & .1694 {\textcolor{gray}{\small $\pm$ .0025}} & \textcolor{blue!55!black}{\underline{.4258 {\textcolor{gray}{\small $\pm$ .0008}}}} & \textcolor{blue!70!black}{\textbf{.1228 {\textcolor{gray}{\small $\pm$ .0002}}}} & \textcolor{blue!55!black}{\underline{1.90}} \\
\bottomrule
\end{tabular}
}
\vspace{-0.1in}
\end{table}

\Tabref{tab:uci} summarizes the results. Overall, DBK performs robustly and consistently across datasets. In terms of MAE, at least one DBK variant attains the best or second-best performance on 6 out of 10 datasets, and this coverage increases to all 10 datasets under NLL and CRPS. Methodologically, DBK-SiLU improves upon VBLL by replacing ELBO-based training with dPPGP, while DBK-RBF similarly improves over SV-DKL and PP-DKL by combining inducing-point bases with the same dPPGP objective. By directly targeting the predictive distribution and incorporating a controllable trace regularizer, dPPGP yields more calibrated uncertainty estimates. MAE shows more modest gains and can occasionally degrade, because objectives that emphasize uncertainty learning (PPGP/dPPGP) may slightly trade off point prediction accuracy. Finally, the variation across datasets highlights the importance of aligning the choice of expansion layers with the data structure.

\Figref{fig:sensitivity} studies the sensitivity of DBK-SiLU to the trace regularization strength $\alpha$ and the KL regularization strength $\beta$ on Pol and Keggdir. The performance benefits from a small but nonzero $\alpha \in [0,1]$, showing the effectiveness of our decoupled trace regularizer. Meanwhile, the performance is less sensitive to the KL weight $\beta \in [0,1]$, consistent with the observation in \citet{jankowiakParametricGaussianProcess2020}.

\subsection{Mobile Internet Quality Estimation}

Finally, we apply the proposed GP regressor to mobile internet quality estimation in Georgia (GA) and New Mexico (NM) using Ookla speed test data\footnote{\url{https://www.speedtest.net/}} \citep{jiangMobileInternetQuality2023}. The dataset consists of quarterly aggregated upload and download speed measurements between 2019 and 2025. The GA and NM datasets contain 517,139 and 133,512 points, respectively, rendering full GP inference intractable. We log-transform the targets to stabilize variance, rescale spatial coordinates to $[-1, 1]^2$, and stratify the $10\%$ test set into urban and rural regions using U.S. Census data\footnote{\url{https://data.census.gov/}}. We compare DBK-RBF trained with dPPGP against RBF and DKL baselines trained with SVGP (\textbf{SV-RBF}/\textbf{SV-DKL}). 

\begin{figure}[t]
  \centering
  \includegraphics[width=0.65\columnwidth]{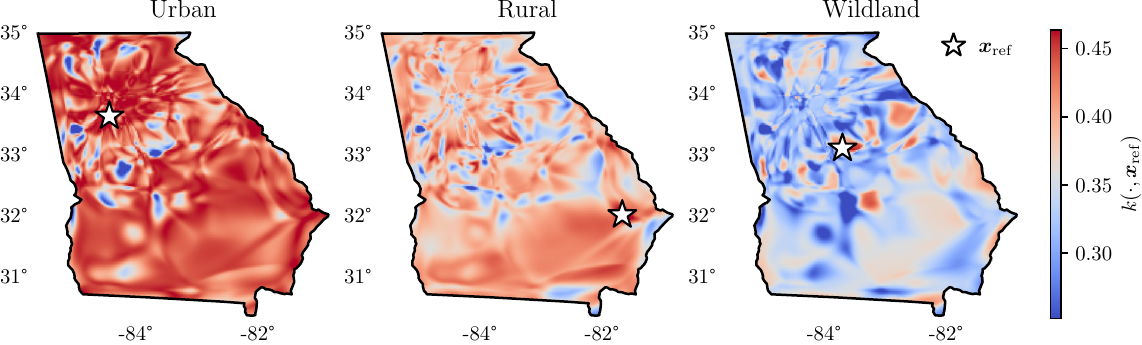}
  \caption{\textbf{Kernel slices $k(\cdot,\vx_{\rm ref})$ at fixed reference locations.} Results are for DBK on GA download speed, showing non-stationary spatial correlations across urban, rural, and wildland regions.}\label{fig:nonstationary}
\end{figure}

\begin{figure}[t]
  \centering
  \begin{subfigure}{0.26\columnwidth}
    \centering
    \includegraphics[width=0.95\textwidth]{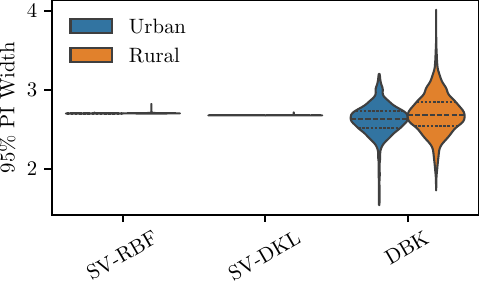}
    \caption{GA, Download Speed}\vspace{.5em}
  \end{subfigure}
  \hspace{.1in}
  \begin{subfigure}{0.26\columnwidth}
    \centering
    \includegraphics[width=0.95\textwidth]{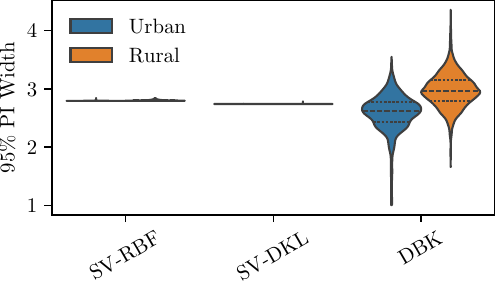}
    \caption{GA, Upload Speed}\vspace{.5em}
  \end{subfigure}
  \\
  \begin{subfigure}{0.26\columnwidth}
    \centering
    \includegraphics[width=0.95\textwidth]{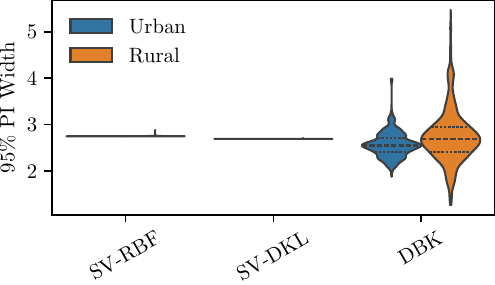}
    \caption{NM, Download Speed}
  \end{subfigure}
  \hspace{.1in}
  \begin{subfigure}{0.26\columnwidth}
    \centering
    \includegraphics[width=0.95\textwidth]{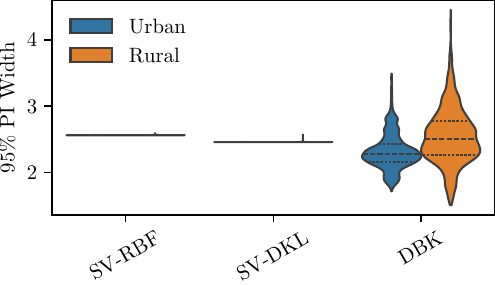}
    \caption{NM, Upload Speed}
  \end{subfigure}
  \caption{\textbf{Violin plots of 95\% PI widths.} Model predictions on the test set, split by Urban/Rural. Additional spatial visualizations of predictive uncertainty are provided in \Figref{fig:ookla_std}.}\label{fig:internet_width}
\end{figure}

As \Tabref{tbl:mobile} shows, DBK consistently attains the lowest NLL across datasets and improves empirical coverage in rural regions. \Figref{fig:internet_width} further highlights qualitative differences in uncertainty estimates: DBK produces meaningful, input-dependent 95\% PI widths, whereas SV-RBF and SV-DKL yield nearly constant intervals across locations, a typical consequence of overestimated noise $\sigma_\eps^2$. For DBK, urban regions exhibit narrower and more concentrated PI widths than rural regions, reflecting stable network conditions in metropolitan areas and high variability in sparsely populated areas. Finally, \Figref{fig:nonstationary} visualizes kernel slices $k(\cdot,\vx_{\rm ref})$ of DBK at representative urban, rural, and wildland locations $\vx_{\rm ref}$ in GA. The kernel shows markedly different spatial patterns at different $\vx_{\rm ref}$. These results illustrate that DBK can simultaneously scale to large datasets, express non-stationary spatial correlations, and deliver well-calibrated uncertainty estimates in real-world settings.

\begin{table}[t]
\centering
\scriptsize
\setlength{\tabcolsep}{3pt}
\renewcommand{\arraystretch}{1.0}
\caption{\textbf{Mobile internet quality estimation.} Test metrics averaged over the test set, split by Urban/Rural. Lower is better for MAE and NLL. The target confidence level is 95\%.}\label{tab:ookla_urban_rural}
\begin{tabular}{@{}>{\raggedright\arraybackslash}p{1.25cm} >{\raggedright\arraybackslash}p{1.25cm} | *{6}{>{\centering\arraybackslash}p{1.05cm}}@{}}
\toprule
\multirow{2}{*}[-.35em]{Dataset} & \multirow{2}{*}[-.35em]{Method} & \multicolumn{3}{c}{Urban} & \multicolumn{3}{c}{Rural} \\
\cmidrule(lr){3-5}\cmidrule(lr){6-8}
 &  & MAE & NLL & Coverage & MAE & NLL & Coverage \\
\midrule
\multirow[t]{3}{*}{GA} & SV-RBF & 0.520 & 1.023 & 96.4\% & 0.544 & 1.074 & 94.5\% \\
Download & SV-DKL & 0.514 & 1.014 & 96.3\% & \textbf{0.531} & 1.055 & 94.7\% \\
 & DBK & \textbf{0.512} & \textbf{1.007} & 96.2\% & 0.531 & \textbf{1.052} & 94.7\% \\
\midrule
\multirow[t]{3}{*}{GA} & SV-RBF & 0.508 & 1.025 & 96.0\% & 0.615 & 1.196 & 93.3\% \\
Upload & SV-DKL & 0.495 & 1.006 & 96.0\% & \textbf{0.593} & 1.164 & 93.3\% \\
 & DBK & \textbf{0.494} & \textbf{0.985} & 95.6\% & 0.603 & \textbf{1.163} & 94.9\% \\
\midrule
\multirow[t]{3}{*}{NM} & SV-RBF & 0.516 & 1.013 & 96.2\% & 0.580 & 1.132 & 93.6\% \\
Download & SV-DKL & 0.507 & 0.998 & 96.1\% & 0.562 & 1.103 & 93.4\% \\
 & DBK & \textbf{0.506} & \textbf{0.989} & 95.3\% & \textbf{0.559} & \textbf{1.076} & 94.6\% \\
\midrule
\multirow[t]{3}{*}{NM} & SV-RBF & 0.456 & 0.921 & 96.4\% & 0.526 & 1.071 & 93.8\% \\
Upload & SV-DKL & \textbf{0.435} & 0.892 & 96.0\% & \textbf{0.495} & 1.011 & 94.1\% \\
 & DBK & 0.437 & \textbf{0.880} & 95.4\% & 0.499 & \textbf{1.002} & 94.6\% \\
\bottomrule
\end{tabular}
\label{tbl:mobile}
\end{table}

\section{Related Works}
\label{sec:lit}


\paragraph{Deep kernel GPs.}
A long line of works integrate NNs into GP models by learning data-dependent representations or kernels. Early approaches apply standard kernels to learned feature spaces \citep{hintonUsingDeepBelief2007,calandraManifoldGaussianProcesses2016}. Deep kernel learning (DKL) \citep{wilsonDeepKernelLearning2016} formalizes this idea by composing a deep NN map with a base kernel, often combined with expressive spectral kernels \citep{wilsonGaussianProcessKernels2013} and scalable inference schemes \citep{wilsonKernelInterpolationScalable2015, wilsonStochasticVariationalDeep2016}. Subsequent work has explored architectural extensions, variational and amortized inference, calibration strategies, and Bayesian treatments to improve scalability or uncertainty quantification
\citep{al-shedivatLearningScalableDeep2017a,tranCalibratingDeepConvolutional2019,daiInterpretableSampleEfficient2020,oberPromisesPitfallsDeep2021,amersfoortFeatureCollapseDeep2022,achituveGuidedDeepKernel2023,liuSimpleApproachImprove2023,matiasAmortizedVariationalDeep2024,zhaoDeepAdditiveKernel2025}.

Another closely related line of works place a Bayesian linear model on top of neural features, yielding a low-rank deep kernel by construction \citep{lazaro-gredillaMarginalizedNeuralNetwork2010,huangScalableGaussianProcess2015}. These include more recent Bayesian last layer (BLL) models \citep{snoekScalableBayesianOptimization2015,kristiadiBeingBayesianEven2020,watsonLatentDerivativeBayesian2021,fiedlerImprovedUncertaintyQuantification2023,harrisonVariationalBayesianLast2024}. Compared to these works, our DBK expands the design space of deep bases and unifies sparse DKL under the same framework. This enables an in-depth analysis of training objectives and uncertainty behavior by generalizing ideas from sparse GPs. 


\paragraph{Low-rank kernel GPs.}
Scalability in GP inference is commonly achieved by exploiting low-rank structure in the covariance. Sparse GP methods approximate the kernel using a small set of inducing variables, leading to Nystr\"om-type approximations and scalable variational objectives
\citep{williamsUsingNystromMethod2000,smolaSparseGreedyMatrix2000,snelsonSparseGaussianProcesses2005,quinonero-candelaUnifyingViewSparse2005,titsiasVariationalLearningInducing2009,hensmanGaussianProcessesBig2013}.
These ideas have been generalized to inter-domain GPs, which define inducing variables by linear functionals 
\citep{lazaro-gredillaInterdomainGaussianProcesses2009,hensmanVariationalFourierFeatures2018,dutordoirSparseGaussianProcesses2020,cunninghamActuallySparseVariational2023}. Other approaches construct low-rank kernels directly using fixed or analytically derived basis functions, including relevance vector machines \citep{tippingSparseBayesianLearning2001}, random and learned Fourier features based on Bochner’s theorem
\citep{rahimiRandomFeaturesLargescale2007,lazaro-gredillaSparseSpectrumGaussian2010,yangCarteLearningFast2015,daoGaussianQuadratureKernel2017,tonSpatialMappingGaussian2018,liTrigonometricQuadratureFourier2024}, and eigenfunctions \citep{wangScalableGaussianProcess2024}.

\section{Conclusions and Future Work}
\label{sec:conc}
We built scalable GPs based on DBKs, a general family of kernels constructed from NN-parameterized basis functions and explicitly enforcing a low-rank structure. We elucidated the design of deep bases and identified the failure modes of standard GP inference under this model class. With these insights, we proposed a mini-batch training objective with decoupled regularization that targets the predictive distribution and prevents rank collapse. Extensive experiments demonstrated the strong performance of DBKs. An important direction for future work is extending DBKs to non-Gaussian likelihoods (e.g., classification). Furthermore, the rank-1 degeneracy we analyzed arises from treating the regression dataset as a single sample path; understanding kernel recovery when multiple sample paths from a ground-truth GP are observed remains an open problem.

\section*{Acknowledgement}

This work is partially supported by an NSF DMS-2134037, CNS-2220387, CMMI-2112533, and the Coca-Cola Foundation.

\bibliography{references}
\bibliographystyle{icml2026}

\newpage
\appendix
\onecolumn

\section{Algorithms}

We summarize the training and prediction procedures for (i) exact MML training of DBK and prediction based on the closed-form posterior, and (ii) mini-batch training of DBK via the proposed dPPGP objective and prediction under the variational formulation.

\renewcommand{\algorithmicrequire}{\textbf{Input:}}
\renewcommand{\algorithmicensure}{\textbf{Output:}}

\begin{algorithm}[h]
\caption{Exact DBK: Training}
\label{alg:exact_dbk_train}
\begin{algorithmic}[1]
\REQUIRE Data $\{(\vx_i,y_i)\}_{i=1}^n$, initial deep basis $\vphi:\R^d\to\R^r$ and noise variance $\sigma_\eps^2>0$
\ENSURE Learned $\vphi(\cdot)$ and $\sigma_\eps^2$
\WHILE{not converged}
    \STATE Compute features $\vphi(\vx_i)$ for $i\in [n]$.
    \STATE Form $\mLambda_\mX \gets \mPhi_\mX^\top \mPhi_\mX + \sigma_\eps^2 \mI_r \in \R^{r\times r}$.
    \STATE Evaluate log marginal likelihood (\ref{eq:lml_dbk}):
    \[
      \log p(\vy) \gets - \frac{n}{2}\log(2\pi) - \frac{n-r}{2}\log (\sigma_\eps^2)  - \frac{1}{2}\log|\mLambda_\mX| - \frac{1}{2\sigma_\eps^2}\|\vy\|_2^2 + \frac{1}{2\sigma_\eps^2}\|\mLambda_\mX^{-1/2}\mPhi_\mX^\top\vy\|_2^2.
    \]
    \STATE Take a gradient step on $\vphi(\cdot)$ and $\sigma_\eps^2$ to maximize $\log p(\vy)$.
\ENDWHILE
\end{algorithmic}
\end{algorithm}
\begin{algorithm}[h]
\caption{Exact DBK: Prediction}
\label{alg:exact_dbk_pred}
\begin{algorithmic}[1]
\REQUIRE Training data $\{(\vx_i,y_i)\}_{i=1}^n$, deep basis $\vphi:\R^d\to\R^r$, noise variance $\sigma_\eps^2>0$, test input $\vx^*$
\ENSURE Predictive mean $\hat\mu_f(\vx^*)$ and variance $(\hat\sigma_f^2(\vx^*)+\sigma_\eps^2)$.
\STATE Compute features $\vphi(\vx_i)$ for $i\in [n]$ and $\vphi(\vx^*) \in \R^r$.
\STATE Form $\mLambda_\mX \gets \mPhi_\mX^\top \mPhi_\mX + \sigma_\eps^2 \mI_r$.
\STATE Compute $\hat\mu_f(\vx^*) \gets \vphi(\vx^*)^\top\mLambda_\mX^{-1}\mPhi_\mX^\top\vy$ and $\hat\sigma_f^2(\vx^*) \gets \sigma_\eps^2\vphi(\vx^*)^\top\mLambda_\mX^{-1}\vphi(\vx^*)$.
\end{algorithmic}
\end{algorithm}
\begin{algorithm}[h]
\caption{DBK via dPPGP: Training}
\label{alg:dbk_dppgp_train}
\begin{algorithmic}[1]
\REQUIRE Data $\{(\vx_i,y_i)\}_{i=1}^n$, initial deep basis $\vphi:\R^d\to\R^r$, noise variance $\sigma_\eps^2>0$, and variational parameters $(\vm,\mL)$, trace regularization weight $\alpha\ge 0$, KL regularization weight $\beta\ge 0$
\ENSURE Learned $\vphi(\cdot)$, $\sigma_\eps^2$, and $(\vm,\mL)$
\WHILE{not converged}
    \STATE Sample a mini-batch $\gB\subset [n]$ of size $b$.
    \STATE Compute features $\vphi(\vx_i)$ for $i\in \gB$.
    \STATE Compute $\hat\mu_f(\vx_i)\gets \langle\vm,\vphi(\vx_i)\rangle$ and $\hat\sigma_f^2(\vx_i)\gets \|\mL^\top\vphi(\vx_i)\|^2$ for $i\in \gB$.
    \STATE Compute trace regularizer $\gL_{\text{trace}} \gets \frac{1}{b} \sum_{i\in\gB} \frac{\tilde{k}_b - \|\vphi(\vx)\|^2}{2\sigma_\eps^2}$ with in-batch proxy $\tilde k_b \gets \max_{i\in\gB} \|\vphi(\vx_i)\|^2$
    \STATE Evaluate mini-batch objective (\ref{eq:dppgp}):
    \[
    \gL_{\text{dPPGP}} \gets \frac{1}{b}\sum_{i\in\gB} - \log \gN (y_i; \hat\mu_f(\vx_i),\hat\sigma_f^2(\vx_i) + \sigma_\eps^2) + \alpha\, \gL_{\text{trace}} + \frac{\beta}{n}\, \KL(\gN(\vm, \mL\mL^\top)\,\|\,\gN(0, \mI_r)).
    \]
    \STATE Take a gradient step on $\vphi(\cdot)$, $\sigma_\eps^2$, and $(\vm,\mL)$ to minimize $\gL_{\text{dPPGP}}$.
\ENDWHILE
\end{algorithmic}
\end{algorithm}
\begin{algorithm}[H]
\caption{DBK via dPPGP: Prediction}
\label{alg:dbk_dppgp_pred}
\begin{algorithmic}[1]
\REQUIRE Deep basis $\vphi:\R^d\to\R^r$, noise variance $\sigma_\eps^2>0$, variational parameters $(\vm,\mL)$, test input $\vx^*$
\ENSURE Predictive mean $\hat\mu_f(\vx^*)$ and variance $(\hat\sigma_f^2(\vx^*)+\sigma_\eps^2)$
\STATE Compute feature $\vphi(\vx^*)$.
\STATE Compute $\hat\mu_f(\vx^*) \gets \langle\vm,\vphi(\vx)\rangle$ and $\hat\sigma_f^2(\vx^*) \gets \|\mL^\top\vphi(\vx^*)\|_2^2$.
\end{algorithmic}
\end{algorithm}

\section{Experiment Details}\label{sec:exp_details}

We implement exact RBF and DKL GP models in GPyTorch \citep{gardnerGPyTorchBlackboxMatrixmatrix2021}, leveraging its highly optimized BBMM algorithms. All other models, including low-rank kernel GPs and DNN, are implemented in PyTorch by ourselves. Unless otherwise stated, we use the dPPGP objective for DBK. All experiments are conducted on a single NVIDIA V100 GPU.

\paragraph{Network architecture.}
We choose $\vg: \R^d \to \R^h$, the backbone of the deep basis map, to be a ResNet \citep{he2016deep,gorishniyRevisitingDeepLearning2023}. The input is first linearly projected from $d$ to $h$ dimensions, after which all residual blocks operate in the $h$-dimensional space. Each residual block applies normalization to the block input, followed by two linear layers with a point-wise nonlinearity in between, and adds the result back to the input via a residual connection. After the residual stack, a final normalization and activation function produce the $h$-dimensional representation. We use SiLU activations, layer normalization, and two residual blocks for all NN-based models and throughout all experiments.

\paragraph{Basis expansion.}
On top of the backbone $\vg$, we consider two choices of expansion layers: 
\begin{enumerate}
    \item \textbf{DBK-SiLU}: The $h$-dimensional backbone output is mapped to $r$ basis functions using a linear layer followed by a SiLU activation. The resulting basis vector is scaled element-wise by a learnable diagonal factor, initialized with random sign flips and scaled by $1/\sqrt{r}$ to stabilize the basis magnitude.
    \item \textbf{DBK-RBF}: We apply an inducing-point kernel expansion to the $h$-dimensional backbone output. The basis is parameterized by a set of learnable inducing points $\mZ \in \R^{r \times h}$ and kernel hyperparameters, including ARD lengthscales $\{\ell_j\}_{j=1}^h$ and a variance $\sigma_k^2$. Inducing points are initialized uniformly in $[-1,1]^h$, and ARD lengthscales are initialized isotropically to $\ell_j=\sqrt{h}$. The RBF kernel with ARD takes the form
    \begin{equation}\label{eq:rbf}
        k(\vx, \vx') = \sigma_k^2 \exp\!\left(-\sum_{j=1}^h \frac{(x_j - x'_j)^2}{2\ell_j^2}\right),
    \end{equation}
    which is used to construct the inducing-point basis as in~(\ref{eq:case_sdkl}).
\end{enumerate}

\paragraph{Variational distribution.}
The variational distribution $q(\vw)$ is parameterized as $\gN(\vm,\mL\mL^\top)$, where $\vm\in \R^r$ and $\mL\in \R^{r\times r}$ is lower triangular with positive diagonals. We represent $\vm$ directly as a learnable parameter. For the Cholesky factor, we store (i) an unconstrained vector for the log-diagonal entries and set the diagonal to $\exp(\cdot)$ to enforce positivity, and (ii) an unconstrained matrix for the off-diagonal entries, from which we keep only the strictly lower-triangular part. We then form $\mL$ by adding the strictly lower-triangular part and the diagonal matrix. We initialize the log-diagonal entries with an offset of $-\tfrac{1}{2}\log r$ and scale the off-diagonal initialization by $1/r$ for numerical stability.

\paragraph{Baselines.}
We compare the proposed DBKs against a range of baselines:
\begin{enumerate}
    \item \textbf{DNN}: A deterministic neural network that directly outputs a prediction using a linear layer atop the backbone $\vg$, trained by minimizing the standard MSE loss.
    
    \item \textbf{RBF} and \textbf{SV-RBF}: Exact GP regression with an RBF kernel and its scalable variant trained with the SVGP objective~(\ref{eq:svgp}). The RBF kernel uses ARD lengthscales as in~(\ref{eq:rbf}). For SV-RBF, inducing points are treated as learnable parameters and initialized uniformly in $[-1,1]^d$. The ARD lengthscales are initialized isotropically to $\ell_j=\sqrt{d}$ for all dimensions, and optimized jointly with the inducing locations and kernel variance under the standard SVGP objective. We adopt the whitened formulation of SVGP \citep{matthewsScalableGaussianProcess2017,jankowiakParametricGaussianProcess2020}, parameterizing the variational distribution over $\vw$ same as described above rather than the inducing variables $\vu$.

    \item \textbf{DKL}, \textbf{SV-DKL}, and \textbf{PP-DKL}: Deep kernel learning \citep{wilsonDeepKernelLearning2016} models that apply an RBF base kernel~(\ref{eq:rbf}) to $h$-dimensional representations extracted by the backbone $\vg$. DKL denotes exact GP inference in the learned latent space. Our SV-DKL baseline treats inducing points as free learnable parameters in the latent space, using the same initialization as DBK-RBF and whitened formulation. Note that this differs from the original SV-DKL paper \citep{wilsonStochasticVariationalDeep2016}, which places inducing points on structured grids and optimizes the GP ELBO using kernel interpolation \citep{wilsonKernelInterpolationScalable2015}. PP-DKL further replaces the SVGP objective with the PPGP objective~(\ref{eq:ppgp}) to improve uncertainty calibration.

    \item \textbf{VBLL}: Variational Bayesian last layer \citep{harrisonVariationalBayesianLast2024}, which places a Gaussian prior on the weights of the final linear layer of an NN and performs variational inference by maximizing the ELBO. For a controlled comparison, we construct VBLL so that the only difference from DBK-SiLU lies in the training objective. Specifically, we use the same basis map $\vphi$ as DBK-SiLU and choose a standard normal prior $\gN(\vzero,\mI_r)$ on the last-layer weights. We do not incorporate additional components from the original VBLL framework, such as structured noise covariance or hierarchical priors. The resulting VBLL objective is
    \begin{equation*}
    \gL_{\text{VBLL}} = \frac{1}{b}\sum_{(\vx, y)} \left\{ 
      - \log \gN (y; \hat\mu_f(\vx), \sigma_\eps^2) + \frac{\hat\sigma_f^2(\vx)}{2\sigma_\eps^2} \right\}
      + \frac{1}{n}\, \KL(\gN(\vm, \mL\mL^\top)\,\|\,\gN(\vzero, \mI_r)).
    \end{equation*}
\end{enumerate}

\paragraph{Hyperparameters.} All models are trained using the AdamW optimizer with a learning rate of $10^{-3}$. Weight decay with coefficient $10^{-2}$ is applied only to the backbone network parameters and not to kernel, variational, or noise parameters. We use a constant mean function, initialized to zero, for all GP models. The noise variance $\sigma_\eps^2$ is initialized to $10^{-2}$ and constrained to be no smaller than $10^{-6}$ for numerical stability. Dataset-specific hyperparameters, such as the hidden size, rank and regularization strengths, are detailed in the corresponding subsections below.

\paragraph{Evaluation protocol.} We evaluate all models using standard metrics for both predictive accuracy and uncertainty quantification. Predictive accuracy is measured by the mean absolute error (MAE), defined as the average absolute difference between predictions and ground-truth targets. Probabilistic performance is assessed using the negative log-likelihood (NLL) under the predictive Gaussian distribution and the continuous ranked probability score (CRPS), which measures the discrepancy between the predictive cumulative distribution function and the empirical distribution. To further assess uncertainty calibration, we report the width of prediction intervals (PI) at a target confidence level of 95\% and the empirical coverage, defined as the fraction of test targets falling within the corresponding predictive intervals. During training, we select the best model checkpoint by monitoring the validation NLL at the end of every epoch. Hyperparameters are tuned based on validation performance. All reported results correspond to the final evaluation on the held-out test set.

\subsection{1-D Synthetic Data}\label{sec:1d_details}

\begin{figure}[t]
  \centering
  \begin{subfigure}[t]{0.47\textwidth}
    \includegraphics[width=\textwidth]{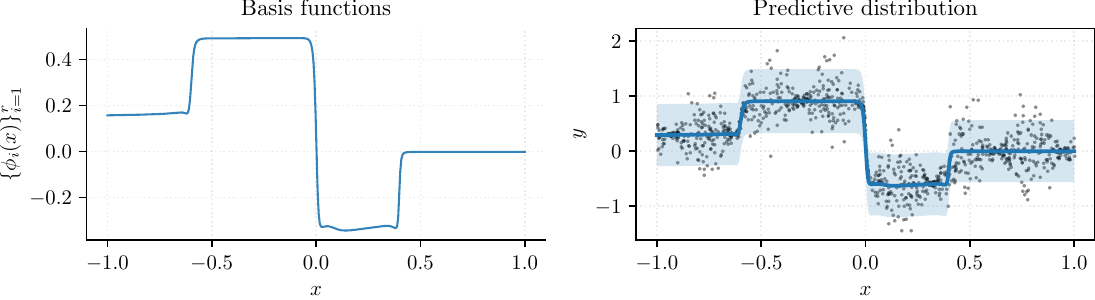}
    \caption{$r = 1$, $\text{Test NLL} = 0.126$}\vspace{.5em}
  \end{subfigure}
  \hfill
  \begin{subfigure}[t]{0.47\textwidth}
    \includegraphics[width=\textwidth]{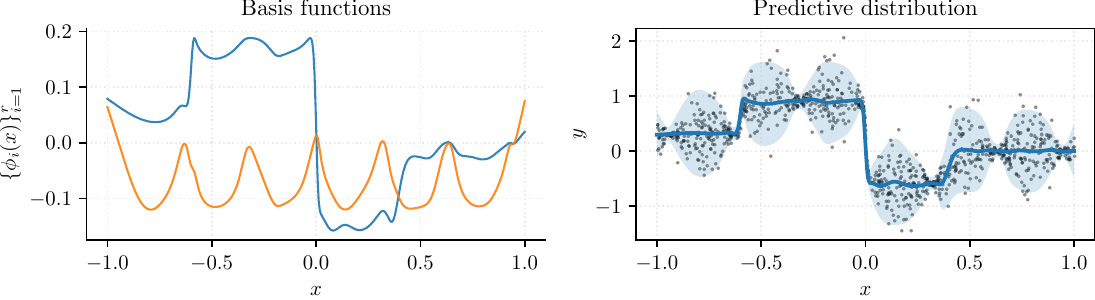}
    \caption{$r = 2$, $\text{Test NLL} = -0.192$}\vspace{.5em}
  \end{subfigure}
  \begin{subfigure}[t]{0.47\textwidth}
    \includegraphics[width=\textwidth]{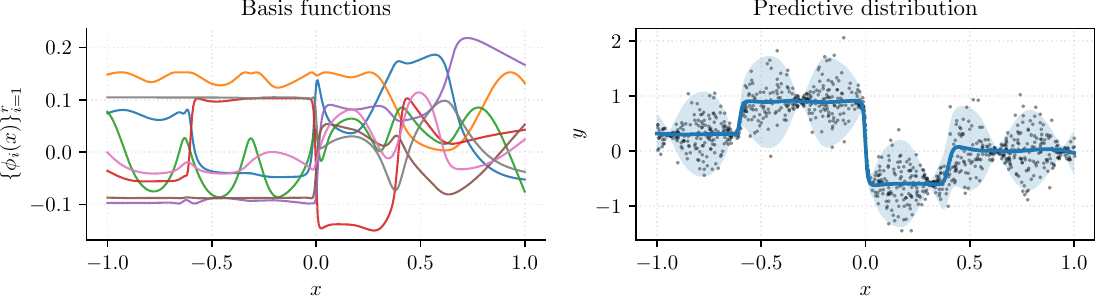}
    \caption{$r = 8$, $\text{Test NLL} = -0.215$}
  \end{subfigure}
  \hfill
  \begin{subfigure}[t]{0.47\textwidth}
    \includegraphics[width=\textwidth]{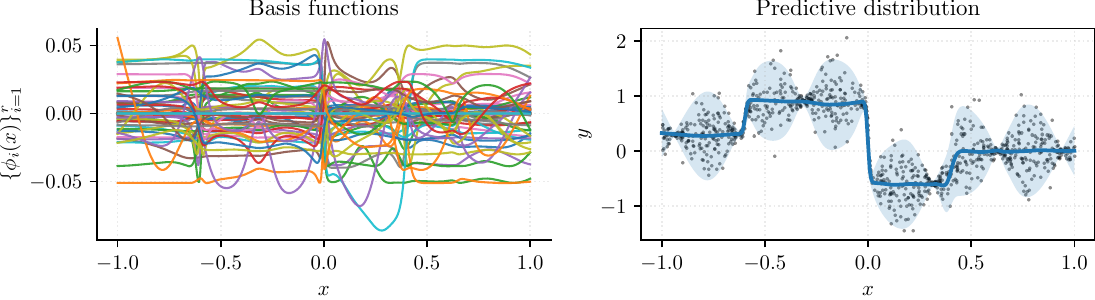}
    \caption{$r = 64$, $\text{Test NLL} = -0.225$}
  \end{subfigure}
  \caption{\textbf{Additional visualizations under different ranks.} We show the basis functions and predictive distributions of DBK (dPPGP) for $r\in \{1, 2, 8, 64\}$ on the 1-D synthetic dataset. For each rank, we show the best-performing run selected by test NLL among 10 random seeds. The experiment setting is identical to that of \Figref{fig:1d_rank}.}
  \label{fig:rank_viz}
\end{figure}

\paragraph{Data generation.} We generate a 1-D regression dataset following the procedure described in Section~\ref{sec:1d}. In total, we sample $100{,}000$ training points, $1{,}000$ validation points, and $1{,}000$ test points. For the scaling experiments, we further construct smaller training sets by randomly subsampling $n \in \{500, 1{,}000, 2{,}000, 5{,}000, 10{,}000, 50{,}000\}$ points from the full training set, while keeping the validation and test sets fixed across all runs.
\paragraph{Hyperparameters.}
The hidden size is set to $h=64$ for the backbone ResNet $\vg$ in DKL and DBK. For dPPGP, we set the trace regularization and KL regularization weights to $\alpha=\beta=0.01$. We set the rank to $r=128$ for Figure~\ref{fig:motivate_exp} and Figures~\ref{fig:step1d_scaling}. To study the effect of rank, we vary the number of basis functions as $r \in \{1, 2, 4, 8, 16, 32, 64\}$, keeping all other hyperparameters fixed. For exact GP models, we perform full-batch training for $2{,}000$ gradient steps, using the modified conjugate gradient algorithm \citep{gardnerGPyTorchBlackboxMatrixmatrix2021} and the Lanczos variance estimates (LOVE) \citep{pleissConstanttimePredictiveDistributions2018}. We use up to $512$ CG iterations and cap the Lanczos rank at $128$.

For DBK trained with dPPGP, we use mini-batch training with batch size $200$ and train for up to $2{,}000$ epochs, employing early stopping when the validation NLL does not improve for $100$ consecutive epochs.

\paragraph{Additional visualizations.} Figure~\ref{fig:rank_viz} provides a qualitative view of how the learned basis functions and predictive distributions evolve as the rank increases. When $r=1$, the model collapses to a single global basis function that resembles the ground-truth mean. As the rank increases to $r=8$, the learned bases become more diverse, enabling the model to represent both sharp transitions in the mean and the sinusoidally varying uncertainty. At $r=64$, the basis functions are highly redundant and distributed, and the predictive distribution closely matches that of lower but sufficient ranks, indicating diminishing returns beyond moderate rank.

\subsection{UCI Regression}

\paragraph{Data preprocessing.}
For each UCI regression dataset, we randomly split the data into training, validation, and test sets with a ratio of $8{:}1{:}1$. The sample size $n$ reported in Table~1 corresponds to the number of training samples. Input features are scaled independently to $[-1,1]$ using min-max normalization computed from the full dataset. Targets are normalized to zero mean and unit variance. The same random split is used across all experiments.

\paragraph{Hyperparameters.}
For all UCI experiments, we set the hidden dimension of the ResNet backbone to $h=64$ and fix the rank to $r=128$ for all low-rank kernels. For DBK trained with dPPGP, we tune the trace and KL regularization weights $\alpha$ and $\beta$ over the grid $\{0, 0.01, 0.1, 1\}$ based on validation NLL. Similarly, the KL regularization weight $\beta$ of PP-DKL is tuned in $\{0, 0.01, 0.1, 1\}$. Models are trained for $400$ epochs with mini-batch stochastic optimization using a batch size of $1024$. We omit standard RBF GPs trained with SVGP or PPGP objectives for the UCI benchmarks, as they consistently underperform neural methods in our preliminary experiments.

\subsection{Mobile Internet Quality Estimation}

\paragraph{Data preprocessing.} The dataset provides quarterly aggregated measurements of mobile internet quality on spatial tiles across the United States. Each record contains the average download or upload speed, the number of tests, and the geographic location of the tile. We focus on two states, Georgia (GA) and New Mexico (NM), and consider download and upload speeds as separate regression targets. For each state, we first filter all records whose spatial locations fall within the corresponding state boundary polygon. The spatial input $\vx\in\R^2$ consists of longitude and latitude coordinates, which are scaled to $[-1,1]^2$ using the bounding box of the state polygon. Targets are log-transformed (base 10) to mitigate skewness and stabilize variance. We treat each quarterly tile measurement as an independent observation, and randomly split the data into training, validation, and test sets with a ratio of $8{:}1{:}1$, using a fixed random seed.

\paragraph{Hyperparameters.} We set the hidden dimension of the ResNet backbone to $h=256$ and fix the rank to $r=1024$ for all methods. For DBK trained with dPPGP, we set the regularization weights $\alpha=0.1$ and $\beta=0.01$. Models are trained for $400$ epochs with mini-batch stochastic optimization using a batch size of $1024$.

\begin{figure}[t]
  \centering
  \begin{subfigure}{0.47\columnwidth}
    \centering
    \includegraphics[width=0.9\textwidth]{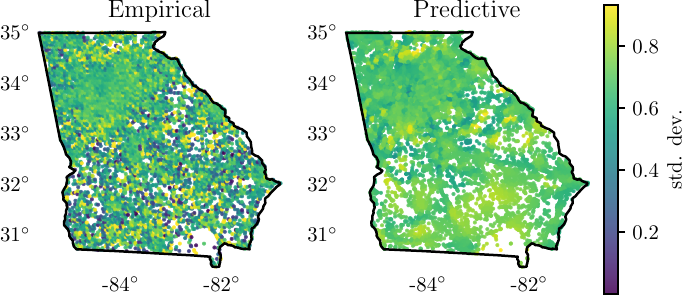}
    \caption{GA, Download Speed}\vspace{.5em}
  \end{subfigure}
  \hfill
  \begin{subfigure}{0.47\columnwidth}
    \centering
    \includegraphics[width=0.9\textwidth]{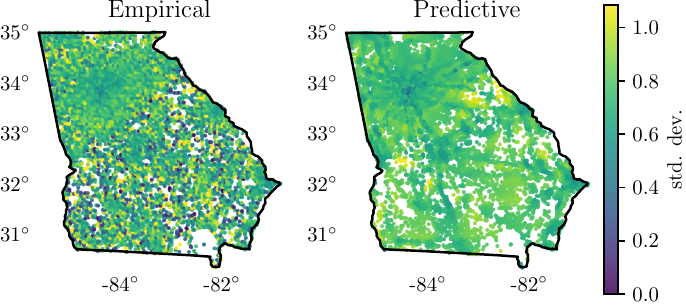}
    \caption{GA, Upload Speed}\vspace{.5em}
  \end{subfigure}
  \\
  \begin{subfigure}{0.47\columnwidth}
    \centering
    \includegraphics[width=0.9\textwidth]{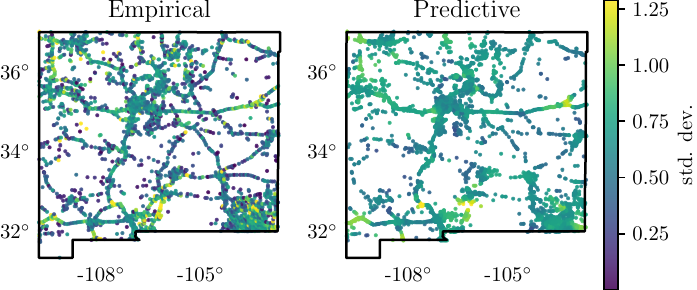}
    \caption{NM, Download Speed}
  \end{subfigure}
  \hfill
  \begin{subfigure}{0.47\columnwidth}
    \centering
    \includegraphics[width=0.9\textwidth]{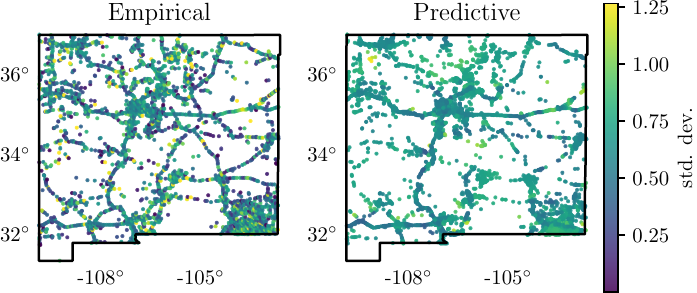}
    \caption{NM, Upload Speed}
  \end{subfigure}
  \caption{\textbf{Spatial comparison of empirical and predictive uncertainty.}
For each state and target, we visualize the empirical standard deviation computed from repeated training measurements aggregated within spatial tiles of size $0.05^\circ \times 0.05^\circ$ (left) and the predictive standard deviation produced by DBK on the test set (right). Both maps share a common color scale, with the range clipped to the 99.5\% percentile to reduce the influence of extreme values.}
  \label{fig:ookla_std}
\end{figure}

\paragraph{Additional visualizations.}
Figure~\ref{fig:ookla_std} provides a spatial comparison between empirical uncertainty estimated from repeated measurements and predictive uncertainty produced by DBK. The empirical standard deviation reflects variability in observed speeds within each spatial tile, while the predictive standard deviation captures the model’s learned input-dependent uncertainty. Across both GA and NM, DBK produces smooth yet spatially varying uncertainty estimates that align well with regions exhibiting higher empirical variability, such as rural and sparsely measured areas. In contrast, more homogeneous urban regions tend to exhibit lower predictive uncertainty. These visualizations qualitatively demonstrate that DBK captures meaningful nonstationary uncertainty patterns consistent with the underlying data heterogeneity.

\section{Proofs and Additional Theoretical Results}

\subsection{Karhunen–Loève Expansion}

\begin{theorem}\label{thm:kl}
    Let $f\in \gX \to \R$ be a zero-mean GP with kernel $k$ satisfying the conditions of Theorem \ref{thm:mercers}. Then, $f$ admits the expansion
    \begin{equation}
        f(\vx) = \lim_{r\to\infty}\sum_{i=1}^r w_i\sqrt{\lambda_i},\psi_i(\vx), \quad w_i \stackrel{\mathrm{iid}}{\sim} \gN(0, 1).
    \end{equation}
    The series converges almost surely.
\end{theorem}
\begin{proof}
Let $\nu$ be a Borel probability measure on compact $\gX$ and define the covariance operator
\[
(\mC g)(\vx)=\int_{\gX} k(\vx,\vx')\,g(\vx')\,d\nu(\vx').
\]
By the assumptions of Theorem \ref{thm:mercers}, $\mC$ is self-adjoint, positive, and trace-class on $L^2(\nu)$ with eigenpairs $\{(\lambda_i,\psi_i)\}_{i\ge1}$ forming an orthonormal basis, and Mercer gives
\[
k(\vx,\vx')=\sum_{i=1}^\infty \lambda_i\psi_i(\vx)\psi_i(\vx')\quad\text{uniformly on $\gX\times\gX$}
\]
Let $w_i\stackrel{\mathrm{iid}}{\sim}\gN(0,1)$ and set $f_r(\vx)=\sum_{i=1}^r \sqrt{\lambda_i}\,w_i\,\psi_i(\vx)$. Then $\E[f_r(\vx)]=0$ and
\[
\Cov (f_r(\vx),f_r(\vx'))=\sum_{i=1}^r \lambda_i\psi_i(\vx)\psi_i(\vx')\to k(\vx,\vx').
\]
Thus for any $\mX = \{\vx\}_{i=1}^n\subset \gX$, $[f_r(\vx_1),\ldots,f_r(\vx_n)]^\top \xrightarrow{d} \gN(\vzero,\mK_{\mX\mX})$, so the limit has the same finite-dimensional distributions as $f$. Moreover $\sum_i\lambda_i=\tr(\mC)=\int_{\gX}k(\vx,\vx)\,d\nu(\vx)<\infty$, hence $\sum_i\lambda_i w_i^2<\infty$ a.s., implying $f_r\to f$ in $L^2(\nu)$ a.s., i.e.,
\[
f(\vx)=\lim_{r\to\infty}\sum_{i=1}^r w_i\sqrt{\lambda_i}\,\psi_i(\vx)\quad\text{a.s.}
\]
\end{proof}

\subsection{Exact DBK via MML}\label{sec:exact}
\paragraph{Log marginal likelihood.}
Denote $\mSigma_\mX=\mPhi_\mX\mPhi_\mX^\top+\sigma_\eps^2\mI_n$ and $\mLambda_\mX=\mPhi_\mX^\top\mPhi_\mX+\sigma_\eps^2\mI_r$. The log marginal likelihood
\[
\log p(\vy)=\log\gN(\vy;\vzero,\mSigma_\mX)
= -\frac{n}{2}\log(2\pi)-\frac{1}{2}\log|\mSigma_\mX|-\frac{1}{2}\vy^\top\mSigma_\mX^{-1}\vy
\]
decomposes into a complexity penalty and a data fit term. By the Sherman--Morrison--Woodbury formula,
\begin{equation}\label{eq:smw}
\mSigma_\mX^{-1}
=\frac{1}{\sigma_\eps^2}\big(\mI_n-\mPhi_\mX\mLambda_\mX^{-1}\mPhi_\mX^\top\big),
\end{equation}
which yields
\[
-\frac{1}{2}\vy^\top\mSigma_\mX^{-1}\vy
=-\frac{1}{2\sigma_\eps^2}\|\vy\|_2^2+\frac{1}{2\sigma_\eps^2}\|\mLambda_\mX^{-1/2}\mPhi_\mX^\top\vy\|_2^2.
\]
By the matrix determinant lemma,
\[
|\mSigma_\mX|=\sigma_\eps^{2(n-r)}|\mLambda_\mX|,
\]
and thus
\[
-\frac{1}{2}\log|\mSigma_\mX|
=-\frac{n-r}{2}\log(\sigma_\eps^2)-\frac{1}{2}\log|\mLambda_\mX|.
\]
Combining terms gives (\ref{eq:lml_dbk}).

\paragraph{Gradient of log marginal likelihood.}
By the chain rule,
\[
\nabla_{\mSigma_\mX}\log p(\vy)
=-\frac{1}{2}\mSigma_\mX^{-1}+\frac{1}{2}\mSigma_\mX^{-1}\vy\vy^\top\mSigma_\mX^{-1},
\]
and
\[
\nabla_{\mPhi_\mX}\log p(\vy)
=2(\nabla_{\mSigma_\mX}\log p(\vy))\mPhi_\mX
=-\mSigma_\mX^{-1}\mPhi_\mX+\mSigma_\mX^{-1}\vy\vy^\top\mSigma_\mX^{-1}\mPhi_\mX.
\]
The required solves are computed via (\ref{eq:smw}):
\[
[\mSigma_\mX^{-1}\mPhi_\mX,\mSigma_\mX^{-1}\vy]
=\frac{1}{\sigma_\eps^2}\Big([\mPhi_\mX,\vy]-\mPhi_\mX\mLambda_\mX^{-1}\mPhi_\mX^\top[\mPhi_\mX,\vy]\Big).
\]

\subsection{Proof of Theorem \ref{thm:oracle_Sigma}}

Under the model $p(\vy)=\gN(\vy;\vzero,\mSigma_{\mX})$ with $\mSigma_{\mX}\succ 0$,
\[
\log p(\vy)
= -\frac{n}{2}\log(2\pi) - \frac12\log|\mSigma_{\mX}| - \frac12\,\vy^\top \mSigma_{\mX}^{-1}\vy.
\]
Taking expectation under $p_{\rm gt}(\vy)=\gN(\vy;\vmu_{\rm gt},\mD_{\rm gt})$ gives
\[
\E_{p_{\rm gt}(\vy)}[\log p(\vy)]
= \text{const} - \frac12\log|\mSigma_{\mX}| - \frac12\,\E_{p_{\rm gt}(\vy)}\!\left[\vy^\top \mSigma_{\mX}^{-1}\vy\right].
\]
Using $\E[\vy^\top \mA \vy]=\tr(\mA\,\E[\vy\vy^\top])$ (for any symmetric $\mA$),
\[
\E_{p_{\rm gt}(\vy)}\!\left[\vy^\top \mSigma_{\mX}^{-1}\vy\right]
= \tr\!\Big(\mSigma_{\mX}^{-1}\,\E_{p_{\rm gt}(\vy)}[\vy\vy^\top]\Big).
\]
Moreover, since $p_{\rm gt}$ is Gaussian,
\[
\E_{p_{\rm gt}(\vy)}[\vy\vy^\top] \;=\; \vmu_{\rm gt}\vmu_{\rm gt}^\top + \mD_{\rm gt}.
\]
Therefore,
\begin{equation}\label{eq:emml}
\E_{p_{\rm gt}(\vy)}[\log p(\vy)]
= \text{const}
-\frac12\log|\mSigma_{\mX}|
-\frac12\,\tr\!\Big(\mSigma_{\mX}^{-1}(\vmu_{\rm gt}\vmu_{\rm gt}^\top+\mD_{\rm gt})\Big).
\end{equation}
Reparameterize by $\mSigma_{\mX}^{-1}\succ 0$. Since $-\log|\mSigma_{\mX}|=\log|\mSigma_{\mX}^{-1}|$,
the objective is a concave function of $\mSigma_{\mX}^{-1}$ (log-det is strictly concave; trace is linear),
so this is a convex optimization problem. The first-order condition is sufficient and necessary.
Differentiating w.r.t.\ $\mSigma_{\mX}^{-1}$ yields
\[
\nabla_{\mSigma_{\mX}^{-1}}\,\E_{p_{\rm gt}(\vy)}[\log p(\vy)]
=\frac12\,\mSigma_{\mX}
-\frac12\big(\vmu_{\rm gt}\vmu_{\rm gt}^\top+\mD_{\rm gt}\big).
\]
Setting the gradient to zero gives the unique maximizer
\[
\mSigma_{\mX}^*
=\vmu_{\rm gt}\vmu_{\rm gt}^\top+\mD_{\rm gt}
=\E_{p_{\rm gt}(\vy)}[\vy\vy^\top].
\]

\subsection{Proof of Theorem \ref{thm:hetero}}

By (\ref{eq:emml}), maximizing $\E_{p_{\rm gt}(\vy)}[\log p(\vy)]$ is equivalent to minimize
\[
\gJ(\mSigma_\mX)=\log|\mSigma_\mX|+\tr(\mSigma_\mX^{-1}\mS_\mX)
\]
over the feasible set
\[
\mSigma_\mX = \mK_{\mX\mX} + \sigma_\eps^2 \mI_n,\qquad \mK_{\mX\mX}\succeq 0,\ \mathrm{rank}(\mK_{\mX\mX})\le r,\ \sigma_\eps^2>0,
\]
Fix any feasible $\mSigma_\mX\succ 0$ and write its eigendecomposition
$\mSigma_\mX = \mV\diag(s_1,\ldots,s_n)\mV^\top$ with $s_1\ge\cdots\ge s_n>0$.
Then $\mSigma_\mX^{-1}=\mV\diag(s_1^{-1},\ldots,s_n^{-1})\mV^\top$.
For fixed eigenvalues $(s_i)$, the term $\log|\mSigma_\mX|=\sum_i\log s_i$ is fixed, and we only need to minimize
$\tr(\mSigma_\mX^{-1}\mS_\mX)$ over $\mV$.
By von Neumann's trace inequality,
the minimum of $\tr(\mSigma_\mX^{-1}\mS_\mX)$ is achieved when $\mSigma_\mX$ and $\mS_\mX$
share eigenvectors $\mV=\mU$. Hence we may restrict w.l.o.g. to $\mSigma_\mX=\mU\diag(s_1,\ldots,s_n)\mU^\top$.

Since $\mK_{\mX\mX}=\mSigma_\mX-\sigma_\eps^2 \mI_n$ has rank at most $r$ and is PSD, $\mSigma_\mX$ must have the form $s_1 \ge \cdots \ge s_r \ge \sigma_\eps^2$ and $s_{r+1} = \cdots = s_n = \sigma_\eps^2$. The objective becomes
\[
\gJ(\mSigma_\mX)
=\sum_{i=1}^r\left(\log s_i+\frac{\lambda_i}{s_i}\right)
+\sum_{i=r+1}^n\left(\log\sigma_\eps^2+\frac{\lambda_i}{\sigma_\eps^2}\right).
\]
For each $i\le r$, the term $\log s_i+\lambda_i/s_i$ as a function of $s_i$ is minimized at $s_i^* = \lambda_i$ on $(0, \infty)$. Plugging $s_i = \lambda_i$, $i=1,\dots,r$ into $\mathcal J$, the only $\sigma_\eps^2$-dependent part is
\[
(n-r)\log \sigma_\eps^2+\frac{1}{\sigma_\eps^2}\sum_{i=r+1}^n\lambda_i,
\]
which has a unique minimizer 
\[
\sigma_{\eps}^{*2}=\frac{1}{n-r}\sum_{i=r+1}^n\lambda_i.
\]
Since $s_i^* \ge \sigma_{\eps}^{*2}$, $i=1,\dots,r$, the solution is feasible. Hence, the optimal kernel matrix is
\begin{align*}
\mK_{\mX\mX}^* & =\mU\diag(\lambda_1,\ldots,\lambda_r,\sigma_{\eps}^{*2},\ldots,\sigma_{\eps}^{*2})\mU^\top-\sigma_{\eps}^{*2}\mI_n \\
& =\sum_{i=1}^r(\lambda_i-\sigma_{\eps}^{*2})\vu_i \vu_i^\top,
\end{align*}

Note that $\mS_\mX=\vmu_{\rm gt}\vmu_{\rm gt}^\top + \mD_{\rm gt}$ is a rank-1 PSD update of
the diagonal matrix $\mD_{\rm gt}$.
Let $d_1\ge\cdots\ge d_n$ be the diagonal entries of $\mD_{\rm gt}$ sorted in decreasing order.
For a rank-1 Hermitian update, eigenvalues interlace \citep{horn2012matrix}:
\[
d_i \ge \lambda_{i+1} \ge d_{i+1},\quad i=1,\ldots,n-1.
\]

\subsection{Derivation of Trace Regularization}\label{sec:trace_derivation}

The trace regularization originally comes from the variational treatment to sparse GPs \cite{titsiasVariationalLearningInducing2009}. Here, we provide a weight-space derivation with a general basis map $\vphi$. 

Assume that the low-rank kernel matrix $\mK_{\mX\mX} = \mPhi_{\mX}\mPhi_{\mX}^\top$ approximates a full kernel matrix $\tilde\mK_{\mX\mX}$ with PSD error $\tilde\mK_{\mX\mX} - \mPhi_{\mX}\mPhi_{\mX}^\top \succ 0$. Given $\vw$, the conditional likelihood under $\mK_{\mX\mX}$ and $\tilde\mK_{\mX\mX}$ can be respectively written as 
\begin{equation*}
  \begin{aligned}
    p(\vy|\vw) & = \gN(\vy; \mPhi_\mX\vw, \sigma_\eps^2\mI_n), \\
    \tilde{p}(\vy|\vw) & = \gN(\vy; \mPhi_\mX\vw, \tilde\mK_{\mX\mX} - \mPhi_{\mX}\mPhi_{\mX}^\top + \sigma_\eps^2\mI_n).
  \end{aligned}
\end{equation*}
We have
\begin{equation*}
  \begin{aligned}
    \log \tilde{p}(\vy|\vw) & \ge \E_{\gN(\vf;\mPhi_\mX\vw,  \tilde\mK_{\mX\mX} - \mPhi_{\mX}\mPhi_{\mX}^\top)}[\log \gN(\vy;\vf, \sigma_\eps^2\mI_n)]                         \\
                            & = \log \gN(\vy;\mPhi_\mX\vw, \sigma_\eps^2\mI_n) - \frac{1}{2\sigma_\eps^2}\tr( \tilde\mK_{\mX\mX} - \mPhi_{\mX}\mPhi_{\mX}^\top) \\
                            & = \log p(\vy|\vw) - \frac{1}{2\sigma_\eps^2}\tr(\tilde\mK_{\mX\mX} - \mPhi_{\mX}\mPhi_{\mX}^\top) \\
                            & = \log p(\vy|\vw) - \sum_{\vx} \frac{\tilde k(\vx,\vx) - \|\vphi(\vx)\|^2}{2\sigma_\eps^2}.
  \end{aligned}
\end{equation*}
The bound is tight when $\tilde{p}(\vf|\vw) = \tilde{p}(\vf|\vw,\vy)$. Marginalizing over $\vw$, we obtain the bound in (\ref{eq:sgp_trace}):
\begin{equation*}
  \begin{aligned}
        \log \tilde{p}(\vy) & = \log \E_{p(\vw)}[\tilde{p}(\vy|\vw)] \\
        & \ge \log \E_{p(\vw)}[p(\vy|\vw)] - \sum_{\vx} \frac{\tilde k(\vx,\vx) - \|\vphi(\vx)\|^2}{2\sigma_\eps^2} \\&  = \log p(\vy) - \sum_{\vx} \frac{\tilde k(\vx,\vx) - \|\vphi(\vx)\|^2}{2\sigma_\eps^2},
  \end{aligned}
\end{equation*}
When the sum is substituted with an in-batch sample average, this gives the trace regularizer:
\[
\gL_{\text{trace}} = \frac{1}{b}\sum_{\vx} \frac{\tilde k(\vx,\vx) - \|\vphi(\vx)\|^2}{2\sigma_\eps^2}.
\]
Note that this term is included in $\gL_{\text{SVGP}}$ (\ref{eq:svgp}) via $\frac{\tilde{\hat\sigma}_f^2(\vx)}{2\sigma_\eps^2}$. For a non-inducing-point basis map $\vphi$, the full kernel $\tilde k$ is typically not defined. To this end, we further substitute $\tilde k(\vx,\vx)$ with an in-batch maximum $\tilde k_b = \max \|\vphi(\vx)\|^2$, resulting in the decoupled trace regularizer in (\ref{eq:dppgp}):
\[
\gL_{\text{trace}} = \frac{1}{b} \sum_{\vx} \frac{\tilde{k}_b - \|\vphi(\vx)\|^2}{2\sigma_\eps^2}.
\]




\end{document}